\documentclass[sigconf]{acmart}
\AtBeginDocument{%
  }
    
\usepackage{multirow}
\usepackage{booktabs, array, tabularx, longtable}
\usepackage[frozencache]{minted}
\usepackage{tcolorbox}
\tcbuselibrary{minted, breakable}
\usepackage{caption}
\usepackage{xspace}
\usepackage{tabularx}
\usepackage{subcaption}
 
\usepackage{pgfplots}                         
\pgfplotsset{compat=1.18}
\usepgfplotslibrary{colormaps}  %

\newcommand{\sysname}{AgentFuel\xspace}

\newcommand{\ignore}[1]{}

\newcommand{\EntitySet}{\mathcal{E}}                      %
\newcommand{\entity}{e}                                   %
\newcommand{\EntityID}[1]{e_{#1}}                         %

\newcommand{\StaticAttrs}{\mathcal{A}}                    %
\newcommand{\staticKey}{a}                                %
\newcommand{\staticVal}[2]{\sigma({#1},{#2})}             %
\newcommand{\StaticProfile}[1]{\mathit{Profile}({#1})} %

\newcommand{\TimeSet}{\mathcal{T}}                        %
\newcommand{\ttime}{t}                                    %
\newcommand{\tstart}{t_{\mathrm{start}}}                  %
\newcommand{\tend}{t_{\mathrm{end}}}                      %
\newcommand{\TimeWindow}[2]{[{#1},{#2}]}                  %

\newcommand{\DynKeys}{\mathcal{K}}                        %
\newcommand{\dynKey}{k}                                   %
\newcommand{\KVPair}[2]{\langle {#1} \mapsto {#2} \rangle}%

\newcommand{\Measurement}[2]{\mathit{Meas}({#1},{#2})}              %
\newcommand{\MeasPayload}[2]{\mathbf{m}({#1},{#2})}       %
\newcommand{\MeasKey}[3]{\mathbf{m}({#1},{#2})[{#3}]}     %
\newcommand{\MeasSet}[1]{\mathcal{M}({#1})}               %

\newcommand{\TimeSeries}[1]{\mathbf{TS}({#1})}             %
\newcommand{\TimeSeriesWindow}[3]{\mathbf{TS}({#1},{#2},{#3})} %

\newcommand{\TimeWin}[3]{W_{\mathrm{time}}({#1},{#2},{#3})}       %
\newcommand{\PredWin}[3]{W_{\mathrm{pred}}({#1},{#2},\phi_{#3})}  %
\newcommand{\KPI}[3]{\mathrm{KPI}_{#1}({#2},{#3})}       %
\newcommand{\KPISet}{\mathcal{F}}                         %

\newcommand{\AttrDist}[2]{\mathbb{P}_{{#1}}^{{#2}}}          %
\newcommand{\AttrDistSpec}[3]{\mathcal{D}({#1},{#2},{#3})}   %
\newcommand{\EntityType}{\mathit{EntityType}}                                 %
\newcommand{\EntityTypeSet}{\mathcal{T}_{\mathrm{ent}}}        %
\newcommand{\AttrSpec}[1]{\mathit{AttrSpec}({#1})}       %

\newcommand{\StateSet}{\mathcal{S}}                            %
\newcommand{\mystate}{s}                                         %
\newcommand{\StateID}[1]{s_{#1}}                               %
\newcommand{\InitState}{s_0}                                   %
\newcommand{\TransFunc}{\delta}                                %
\newcommand{\TransProb}[3]{p({#2} \to {#3} \mid {#1})}        %
\newcommand{\StateDuration}[2]{\Delta({#1},{#2})}              %
\newcommand{\STD}[1]{\mathit{StateTrans}({#1})}                       %
\newcommand{\ActiveState}[2]{\mystate({#1},{#2})}                %

\newcommand{\MeasSpec}[2]{\mathit{MeasSpec}({#1},{#2})}                    %
\newcommand{\ValDomain}[1]{D_{#1}}                             %
\newcommand{\MeasGenFn}[3]{\hat{\mu}({#1},{#2},{#3})}          %

\newcommand{\Exemplar}[1]{\mathit{Exemplar}_{{#1}}}                  %
\newcommand{\ExemplarSet}{\mathit{ExemplarSet}}                           %
\newcommand{\ExemplarBehavior}[1]{\mathit{Behavior}({#1})}                 %
\newcommand{\ExemplarEntities}[1]{N_{{#1}}}                    %
\newcommand{\ExemplarDuration}[1]{T_{{#1}}}                    %

\newcommand{\Chunk}[1]{\mathit{epoch}_{{#1}}}                               %
\newcommand{\ChunkWindow}[1]{\TimeWindow{t_{{#1}}}{t_{{#1}+1}}} %
\newcommand{\BlendWeight}[2]{w_{{#1}}^{{#2}}}                   %
\newcommand{\BlendFn}[2]{\mathcal{B}({#1},{#2})}               %
\newcommand{\GlobalDataset}{\mathbf{D}_{\mathrm{global}}}      %
\newcommand{\ChunkDataset}[1]{\mathbf{D}({#1})}                %
\newcommand{\BlendProfile}[1]{\mathbf{w}({#1})}                %

\newcommand{\EventType}[1]{\mathtt{#1}}
\newcommand{\EventPred}[1]{\mathit{EventFilter}^{\mathrm{ev}}_{#1}}
\newcommand{\EventPredSet}{\mathit{EFSet}}
\newcommand{\EventSeq}[1]{\langle #1 \rangle}

\newcommand{\QueryTemplate}[1]{\mathbf{Q}_{#1}}
\newcommand{\QueryResult}[1]{\mathbf{R}_{#1}}
\newcommand{\QueryWindow}{W}
\newcommand{\EntityFilter}[1]{\mathit{EntityFilter}_{\mathrm{ent}}(#1)}
\newcommand{\GroupBy}[2]{\mathrm{GROUP}_{#1}({#2})}

\newcommand{\Exists}[1]{\mathrm{EXISTS}(#1)}
\newcommand{\Count}[1]{\mathrm{COUNT}(#1)}
\newcommand{\AvgTimeBetween}[2]{\mathrm{AVG\_TIME}(#1 \to #2)}
\newcommand{\SeqMatch}[1]{\mathrm{SEQ\_MATCH}(#1)}
\newcommand{\ConvRate}[2]{\mathrm{CONV}(#1 \to #2)}
\newcommand{\Agg}[2]{\mathrm{#1}({#2})}
\newcommand{\Percentile}[2]{P_{#1}({#2})}
\newcommand{\CompareWindows}[3]{\Delta_{#1}({#2},{#3})}

\newcommand{\AltCount}[1]{\mathrm{ALT\_COUNT}(#1)}

\newcommand{\WinMean}[3]{\bar{#1}({#2},{#3})}
\newcommand{\WinStd}[3]{\hat{\sigma}[{#1}]({#2},{#3})}
\newcommand{\WinCount}[2]{N({#1},{#2})}
\newcommand{\WinRate}[3]{\rho_{#1}({#2},{#3})}

\newcommand{\StateEntry}[1]{\mathrm{entry}(#1)}
\newcommand{\StateExit}[1]{\mathrm{exit}(#1)}
\newcommand{\StateTimeout}[1]{\tau_{#1}}
\newcommand{\StateOcc}[3]{\mathrm{OCC}({#1},{#2},{#3})}
\newcommand{\StateDur}[3]{\mathrm{DUR}({#1},{#2},{#3})}
\newcommand{\StateReached}[3]{\mathrm{STATE\_REACHED}({#1},{#2},{#3})}
\newcommand{\CountInState}[4]{\mathrm{CIS}({#1},{#2},{#3},{#4})}
\newcommand{\TransMatrix}[2]{\mathbf{T}({#1},{#2})}
\newcommand{\CommonPaths}[2]{\mathrm{PATHS}({#1},{#2})}
\newcommand{\LoopCount}[3]{\mathrm{LOOPS}({#1},{#2},{#3})}
\newcommand{\OccupancyDist}[3]{\boldsymbol{\pi}({#1},{#2},{#3})}
\newcommand{\FirstPassage}[3]{\mathrm{FP}({#1},{#2},{#3})}

\newcounter{packednmbr}
\newenvironment{packedenumerate}{\begin{list}{\thepackednmbr.}{\usecounter{packednmbr}\setlength{\itemsep}{0.5pt}\addtolength{\labelwidth}{-15pt}\setlength{\leftmargin}{\labelwidth}\setlength{\listparindent}{\parindent}\setlength{\parsep}{1pt}\setlength{\topsep}{0pt}}}{\end{list}}

\newenvironment{packeditemize}{\begin{list}{$\bullet$}{\setlength{\itemsep}{1pt}\addtolength{\labelwidth}{0pt}\setlength{\leftmargin}{\labelwidth}\setlength{\listparindent}{\parindent}\setlength{\parsep}{1pt}\setlength{\topsep}{0pt}}}{\end{list}}

\newcommand{\myparagraph}[1]{\smallskip \noindent{\bf {#1}:}~}

\setcopyright{none}

\begin{document}

\title{Generating Expressive  and Customizable Evals for \\ Timeseries Data Analysis Agents with \sysname}

\author{Aadyaa Maddi, Prakhar Naval,   Deepti Mande, Shane Duan, Muckai Girish, Vyas Sekar \\ Rockfish Data and Carnegie Mellon University \\ 
Feb 28 2026, Pre-print}

\pagestyle{plain}

\begin{abstract}
Across many domains (e.g., IoT, observability, telecommunications, cybersecurity), there is an emerging adoption of {\em conversational data analysis agents} that enable users to ``talk to your data'' to extract insights. Such  {\em data analysis  agents} operate on {\em timeseries data models}; e.g., measurements from sensors or events monitoring user clicks and actions in product analytics. 
We evaluate 6 popular data analysis agents (both open-source and proprietary) on domain-specific data and query types, and find that they fail on stateful and incident-specific queries. We observe two key expressivity gaps in existing evals: domain-customized datasets and domain-specific query types.  
To enable practitioners in such domains to generate customized and expressive evals for such timeseries data agents, we present \sysname. \sysname helps domain experts quickly create customized evals to perform end-to-end functional tests. We show that \sysname's benchmarks expose key directions for improvement in existing data agent frameworks. We also present anecdotal evidence that using \sysname can improve agent performance (e.g., with GEPA). \sysname benchmarks are available at \url{https://huggingface.co/datasets/RockfishData/TimeSeriesAgentEvals}.

\end{abstract}

\maketitle

\section{Introduction}

Across many industries, we are  seeing the adoption of  data analysis agents (e.g.,~\cite{databricksgenie,grafanaassistant,hextech,openaiagent2024,elasticassistant,mixpanelassistant,thousandeyesassistant}). This includes their adoption in product analytics domains, observability domains,  cybersecurity domains  as well as more broadly (e.g.,~\cite{forrester2026agents,polystar2026mwc,google2026vertex,digitalgenius2025ecommerce,redcanary2025soc,mckinsey2025trends}. %
In contrast to traditional dashboards and code-based analysis or notebooks,  such  data  agents can democratize access to deep data, driven insights beyond expert data scientists, and coders. 
There have been significant advances in various aspects of the data analytics pipelines. This includes work on advanced Text2SQL benchmarks (e.g.,~\cite{lei2024spider2,bird,condabench2025}), improving the data pipelines~\cite{tapilot2024},  techniques for including more enterprise data context systematically (e.g.,~\cite{openaiagent2024,baek2025knowledge,chhikara2025mem0}),  as 
well as practical implementations being rolled out 
 in pilots and production. %

Our specific focus in this paper is on data analysis agents that deal with {\em enterprise time series data}. These appear in many popular enterprise domains, including telecommunications, observability, cyber security, IoT,  manufacturing, critical infrastructure utilities, and so on~\cite{grafanaassistant,hextech,openaiagent2024,elasticassistant,mixpanelassistant,thousandeyesassistant}. These domains routinely collect a wide range of telemetry data, either capturing key metrics of interest from agents in the wild or measure user activity in the form of like events occurring during a user interaction with their applications.

 A natural concern for practitioners with the deployment of such conversational data agents as with any other agent workflows in general, is their {\em reliability}~\cite{cemri2025multiagentllmsystemsfail,rabanser2026towards,zhu2025survey}.  That is, do the responses and answers provided by the agent when consumers and clients talk to the data correctly represent the expected answers.

 This paper is motivated by a simple question of helping   practitioners   deploying agents over time series data in specific domains such as monitoring, IoT, and observability.
 More specifically, in our settings practitioners are often interested in: (1)  {\em stateful  contextual} analytics  that capture sequence and timing effects; e.g., did a user abandon the cart within 10 minutes after adding three or more items  and (2) {\em anomalous  incidents} to capture patterns in data that seem out of the ordinary; e.g.,  did the host send a lot of outbound traffic after downloading a malicious file.

We evaluate  many state-of-art agents that do  well on traditional  basic queries in benchmarks and public datasets, but find that they produce incorrect and inconsistent answers on domain-specific  data  and query patterns of interest.

There is an imminent need for adopting an evaluation-driven development framework for the design, implementation and deployment of   data analysis agents~\cite{anthropic2026demystifying}. 
 In our  context, we see that  current evals~\cite{lei2024spider2,li2023llmservedatabaseinterface}  have  two key {\em expressivity gaps}.  First, we find a  {\em  dataset gap} to  capture domain-specific data semantics of interest both in terms of ``normal'' and ``unexpected'' patterns. For instance, in observability or security, the  datasets should capture the  semantics  of diverse normal and incident-specific pattern structures. 
 Second, we find an {\em analytic queries gap}  where agents posed with  questions and personas  of relevance to the domain falter.\footnote{To be fair, existing benchmarks do expose other kinds of complementary issues with  data agents; our  focus here is on the kinds of data and query patterns relevant to timeseries data agents in specific domains.}

In this paper, we present the design and implementation of {\em \sysname} -- an expressive and customizable {\em evals framework}. \sysname  enables practitioners to quickly create customized evals  for testing their time series data analytics agents before they rolled out in production.  We  focus on end-to end   functional evaluation of the agent capabilities~\cite{anthropic2026demystifying} to complement other mechanisms   that look at  reasoning  steps, tool calls, or backend request traces~\cite{arize,langchain}.

Given a high-level description  of 
 the data source (e.g., schema or sample data) and the agent to be tested, \sysname   
 produces  a {\em benchmark} consisting of 
  a set of reference datasets  and attendant 
   query-answer pairs.  \sysname  consists of two interconnected steps   to create customized evals for timeseries analytics agents.   The first step  is an AI-assisted workflow for generating time series datasets   customizable to the deployment scenario of interest capturing data model semantics 
    and  adding domain-specific interesting incident  patterns of interest.  
    Given the generated dataset, the second step creates a set of curated question-answer pairs that is {\em coupled } to the dataset and domain; e.g., asking questions about specific incident patterns of interest  and simulating variants of questions posed by user personas of relevance to the domain.

We implement \sysname as a modular extensible  pipeline.  
We develop a proprietary LLM-assisted  data generation pipeline that allows us to leverage the ``world model'' of the LLM together with the controllable generation of structured pipeline using a Python SDK. We develop an extensible and controllable ``pattern injection'' library that covers many common patterns that are common in the enterprise domains of interest (e.g., KPI degradation, data outages, sudden flash crowds).    We also develop an extensible library of SQL-based data quality and data pattern checkers  to ensure that the generated data  meets the scenarios that the expert intended. 

Along with the data quality checks, we also develop a set of structured analytical queries {\em coupled} to the data and injected patterns of interest; e.g., if there was a spike in a timeseries of a metric, we include queries that check for the spike or detect the spike or analyze other KPIs in the duration around the spike. Finally, we use a structured prompt variant generator to go from structured queries to natural language variant~\cite{sql2nl2025}.

To evaluate \sysname,  we create a number of domain-specific datasets of interest in three different timeseries settings: product analytics for e-commerce, IoT sensor data, and telecom radio access network (RAN) data. We evaluate popular   industry and open-source agents, such as Databricks  Genie~\cite{databricksgenie}, Snowflake Cortex~\cite{snowflakeanalyst}, PandasAI~\cite{pandasai_repo}, Nao~\cite{nao_repo} following best practices. 
We find that even though these agents can achieve 73\% accuracy on simple aggregation type queries, their accuracy on the stateful and incident-specific queries produced by \sysname is substantially lower in the 10\% range. Furthermore, we show how preliminary evidence that using the \sysname  evals in a GEPA~\cite{agrawal2026gepareflectivepromptevolution} optimization loop can help improve the accuracy by 17\%.

\section{Background and Motivation}

\label{sec:motivation}

We begin by describing   motivating scenarios of  enterprises building ``talk to your data''   agents over time series data.  We identify requirements from these motivating scenarios and highlight gaps in existing evaluation frameworks. 
\begin{figure}[th]
    \centering
    \includegraphics[width=0.95\linewidth]{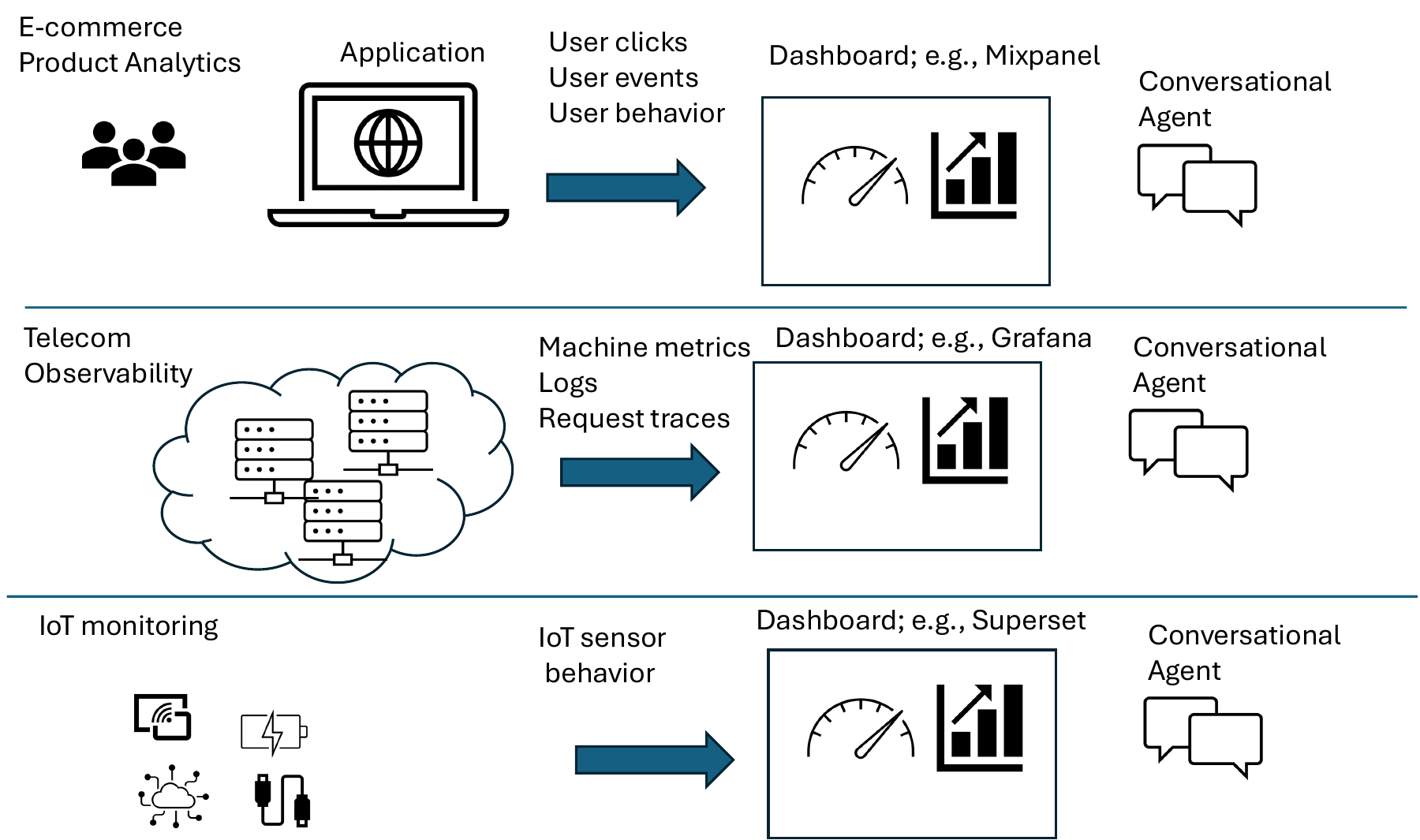}
       \vspace{-0.3cm}
    \caption{Setting: Many domains are  building a ``talk to your data'' agent for conversational data analytics}     \vspace{-0.3cm}
    \label{fig:placeholder:motivation}
\end{figure}

\begin{table*}
	[th]
	\begin{footnotesize}
		\begin{tabular}{l|p{3.5cm}|p{2cm}|p{2.5cm}|p{2.5cm}|p{2.5cm}}
			Setting  & Query & \multicolumn{4}{c}{Data Agents}\\
			\cline{3-6} & & Databricks Genie & Snowflake Cortex Analyst & Nao GPT 4.1 & PandasAI Opus 4.6 \\
			\hline
			\multirow{3}{*}{E-commerce product analytics} & What is the average time between viewing a product and adding it to cart? & Inconsistent answers across independent runs & Incorrect answer, no state tracking & Inconsistent answers across independent runs & Inconsistent answers across independent runs \\
			\cline{2-6} & How long on average do users spend with items in cart before purchasing or abandoning? & Incorrect answer, no state tracking & Incorrect answer, runtime error & Incorrect answer, no state tracking & Inconsistent answers across independent runs \\
			\hline
			\multirow{3}{*}{IoT sensor measurements} & What is the average time between threshold exceeded and maintenance required? & Incorrect answer & Incorrect answer & Inconsistent answers across independent runs & Correct answer  \\
			\cline{2-6} & How many readings were recorded while sensors were being maintained? Show me a breakdown by device type & Incorrect answer, no state tracking & Incorrect answer, no state tracking & Incorrect answer, no state tracking & Incorrect answer, no state tracking \\
			\hline
			\multirow{3}{*}{Telecom observability} & What was the average latency on the affected router link during the incident? & Incorrect answer, no incident detection & Incorrect answer, no incident detection & Incorrect answer, runtime error & Incorrect answer, no incident detection \\
			\cline{2-6} & How many cells lost availability during the outage on January 2, while the core nodes were also under load? & Incorrect answer, no incident detection & Incorrect answer, no incident detection & Incorrect answer, no incident detection & Incorrect answer, no incident detection \\
			\hline
		\end{tabular}
	\end{footnotesize}
	\caption{
    State-of-the-art data analysis agents mostly fail on natural timeseries queries. We evaluate four agents -- Databricks Genie, Snowflake Cortex Analyst, Nao GPT4.1, and PandasAI Opus 4.6 -- on three timeseries datasets from relevant application domains (e-commerce, IoT, telecom). 
    Appendices ~\ref{sec:appendix:morefail} and ~\ref{sec:appendix:listings} show more examples.}
	\label{tab:sec2:summary}
\end{table*}

\subsection{Motivating Scenarios}

\myparagraph{Product analytics}  Many providers offer product analytics services (e.g.,~\cite{mixpanelassistant}). They track the  user behaviors (e.g.  clicks, actions, search queries)  on their websites.  On top of these  event data,  they offer a wide range of analytics  to help their customers identify issues and improvement opportunities.  For example, a dashboard might show the number of customers that  abandoned products or   a breakdown of the page response times for different types of devices.

\myparagraph{Telecom and observability} Observability (O11Y) is the act of collecting telemetry from backend systems serving applications~\cite{elasticassistant,datadog,thousandeyesassistant}. 
 For instance, these collect timeseries data of metrics and log events to   check if there are problems and  troubleshoot when performance  problems arise. For example, did the user-perceived latency go up when assigned to high CPU load servers?

\myparagraph{IoT   monitoring}  Similar to the above settings, IoT systems collect timeseries metrics and events from sensors in the field (e.g.,~\cite{awsiot}). The sensor data is fed into dashboards and databases to enable further analytics. For instance, operators may want to analyze if the sensors were raising warning signals and for how long to trigger predictive maintenance workflows.

Classical timeseries analytics workflows in these domains\footnote{We defer a more formal definition of the data and query model until Section~\ref{sec:design}.}  involved dashboards and human operators.  As with many other domains, providers  are deploying  {\em data analysis agents} to help users  get insights faster through a conversational frontend.
 While there is significant  excitement to lower the barrier  to get useful   insights, there is a significant reliability bottleneck as we will see next.

 \subsection{How well do  data agents do in our  setting?} %

A natural question  is how well do existing data agents fare in these specific settings.   In particular,  domain practitioners in such timeseries settings are interested in {\em temporal behaviors of interest}; e.g., did a metric spike, did a user do a specific sequence of actions, how long did some event last, and so on. 

To  this end, we created  illustrative synthetic datasets for three domains of interest and illustrative queries. We took 4 popular industry and open-source  data analytics agents and tested how well they fare for illustrative queries.  We defer  more details on the setup to Section~\ref{sec:evaluation}. 
 Note that these agents perform  well in classical Text2SQL benchmarks (e.g.,~\cite{cortexbenchmark,geniebenchmark,naobenchmark}).  
 Our goal here is not to highlight shortcomings of these  agents  but to highlight a broader {\em expressivity gap} in evals for timeseries agents.

 Table~\ref{tab:sec2:summary} shows  illustrative queries for three  domains -- product analytics, telecom, and IoT monitoring. We see that   agents perform poorly and produce inconsistent or incorrect answers. To put this in context, we also asked some more generic (non temporal)  queries~\cite{lei2024spider2,bird} and found that the  agents did well (not shown). That is, the issue is not that the agents are bad, but that there is a specific {\em expressivity} gap  for our domain-specific queries and intents.

\subsection{ Gaps in Existing Evals}

\begin{figure}[t]
  \centering  
  \begin{tikzpicture}
	\begin{axis}[
		ybar,
		bar width=14pt,
		width=\columnwidth,
		height=5cm,
		symbolic x coords={{Spider2-Snow},{BIRD},{Beaver}},
		xtick=data,
		xticklabel style={rotate=0, anchor=north, font=\small},
		enlarge x limits={abs=1.5cm},
		clip=false,
		ymin=0,
		ymax=100,
		ylabel={Percentage (\%)},
		legend style={at={(0.5, 1.12)}, anchor=south, font=\small, legend columns=3},
		ymajorgrids=true,
		grid style=dashed,
	]
		\addplot[fill=teal!60] coordinates { ({Spider2-Snow}, 92.32) ({BIRD}, 96.33) ({Beaver}, 93.78) };
		\addlegendentry{Stateless}

		\addplot[fill=violet!70] coordinates
		{ ({Spider2-Snow}, 6.95) ({BIRD}, 3) ({Beaver}, 6.22) }; \addlegendentry{Stateful, No Incident}

		\addplot[fill=brown!55] coordinates
		{ ({Spider2-Snow}, 0.73) ({BIRD}, 0.67) ({Beaver}, 0) }; \addlegendentry{Incident}
	\end{axis}
\end{tikzpicture}
  \vspace{-0.4cm}
  \caption{Queries in SOTA benchmarks are mostly ``stateless''}
    \vspace{-0.4cm}
  \label{fig:fig0_sota_benchmark_queries}
\end{figure}
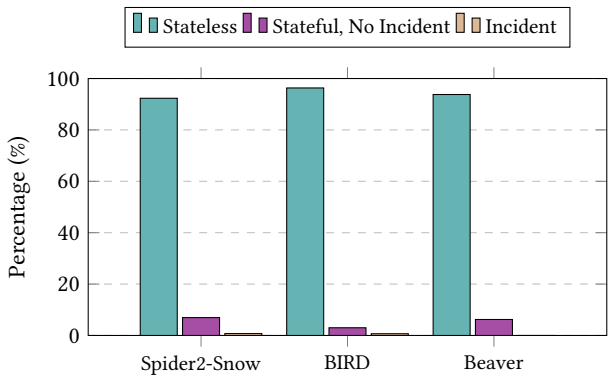

 As the authors in ~\cite{cortexbenchmark} observe, time-series data, like sales or user activity, is often underrepresented in traditional benchmarks -- a gap our evaluation confirms. We analyze three widely-used benchmarks: \textsc{Spider2-Snow} ~\cite{lei2024spider2}, \textsc{BIRD LiveSQLBench-Base-Full-v1}~\cite{bird}, and \textsc{Beaver}~\cite{chen2024beaver}. We classify each query in the benchmark into one of three categories: stateless, stateful (no incident), or incident. We apply a keyword-based heuristic classifier to the ``gold'' SQL query when one is available, and fall back to classifying the natural language question otherwise. Figure~\ref{fig:fig0_sota_benchmark_queries} shows  the  majority of queries in existing benchmarks are stateless (92\% in Spider2Snow, 96\% in BIRD LiveSQLBench-Base-Full-v1, 94\% in Beaver). These queries require agents to do a single lookup or aggregation over a fixed snapshot. They do not test agents for stateful, incident-aware reasoning that practitioners require.
We identify two key  gaps in current benchmarks:
\begin{itemize}
    \item {\em  Dataset expressivity:} Existing  evals focus on generic  or public data. 
    As such, they  do not provide semantically relevant data for a specific domain.  Further,  practitioners would want eval ``ground truth'' datasets to   interesting events or incidents or anomalies (e.g., spikes in metrics, changes in user behavior).   
    \item {\em  Query expressivity:} In our setting,  practitioners want to ask  stateful  contextual analysis  pertaining  to normal behaviors (e.g., what happened to KPI when CPU was high) and   incidents of interest. Further, they may have different domain specific personas or linguistic patterns of interest.  Our illustrative queries show  that many  agents stumble on stateful and incident-related queries. 
    
\end{itemize}

\section{\sysname System Design}
\label{sec:design}
 Next, we discuss the design of \sysname. We start with a high-level overview  and problem scope before describing the detailed design. 
\subsection{Overview}
\begin{figure}[t]
    \centering
    \includegraphics[width=0.99\linewidth]{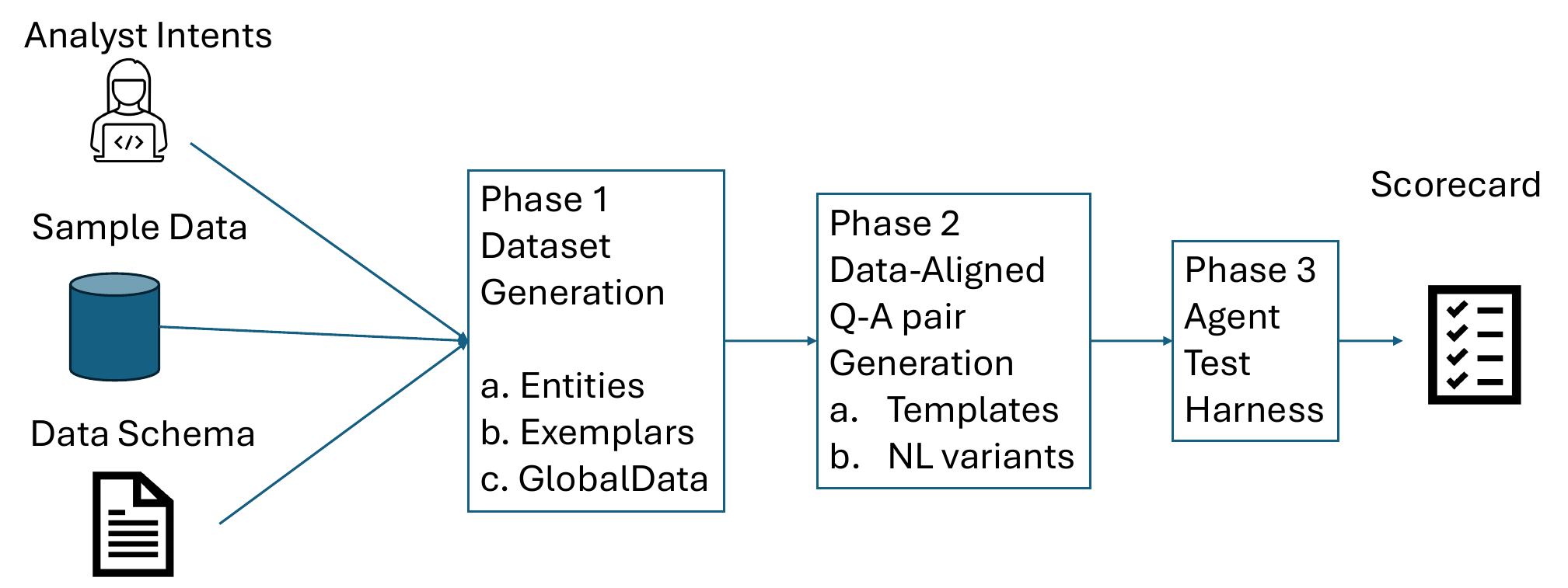}
    \vspace{-0.2cm}
    \caption{Overview of the proposed \sysname system with key modules to tackle the requirements }
       \vspace{-0.2cm}
    \label{fig:overview}
\end{figure}

Our goal in designing \sysname  is to lower the effort  to create bespoke  {\em expressive and customized evals}  for   time series data agents.
Our approach is guided by three key design tenets:
\begin{packeditemize}

\item {\em End-to-end deterministic  evals:} 
First, we focus on end-to-end functional evaluations of the data agent rather than focus on intermediate steps (e.g., code   or tool calls).    With the rapid evolution of compound agent systems~\cite{compoundai} and constant evolution of models and best practices, and the  ability of agents to dynamically generate code, we believe that such a functional  evaluation is necessary to complement 
 other criteria that  look at   traces, tool calls, or  SQL queries generated.

    \item {\em Representative data:} Second, we argue the evals to be representative of the data  patterns that appear in the domain of interest. For instance, if we are evaluating a IoT or Observability data agent, then the dataset to evaluate the agent must capture natural semantics (e.g., seasonality, heterogeneity). 
    
    \item {\em Data-query alignment:} Third, we  argue for  explicitly ensuring  data and query alignment. To see why, consider the  following example. Suppose the query checks if  there is a spike in some user relevant KPI of interest.  If  we  have a naïve agent that always responds   that there are no spikes. Now, if the eval data does not actually contain such incidents, we would incorrectly conclude that the agent was working correctly.
    
\end{packeditemize}  

\myparagraph{Scope} We scope this paper  on two fronts. First, we focus  on single-turn interactions  and do not include multi-turn conversations or historical context. %
We leave multi-turn interaction for future work. Second, we  focus  on testing timeseries agents before they are deployed. 
 That  said, we do believe that  \sysname  could also help in other parts of the agent development lifecycle; e.g.,  training data set or continuous improvements.

\myparagraph{Workflow}
Figure~\ref{fig:overview}  shows a high-level overview of \sysname. It  takes as input a data schema or sample data from the domain expert. \sysname can also take in other optional inputs or hints that the domain expert wants to provide for guiding the design of the evals. For instance, the practitioner could specify the kinds of anomalies or incidents, query patterns, or personas relevant to the domain. \sysname works in three logical phases as shown in Figure~\ref{fig:overview}: dataset generation, question-answer generation, and test integration. We describe the detailed design of each next.

\begin{table}[th]
\begin{footnotesize}
\begin{tabular}{p{2cm}| p{1.5cm}| p{3.5cm}}
\textbf{Term} & \textbf{Symbol} & \textbf{Description / Example} \\ \hline
Entity & $\EntityID{}$ & Some object  under observation (e.g., users, sensors, or hosts) \\
Static Attributes & $\StaticAttrs$ & Fixed characteristics such as OS version, location, or service type. \\
Dynamic Keys & $\DynKeys$ & Set of dynamic attribute keys (e.g., CPU, memory). \\
Measurement & $\Measurement{\entity}{\ttime}$ & Set of observed key-value pairs for entity $\entity$ at time $\ttime$. \\
Time Series & $\TimeSeries{\entity}$ & Chronologically ordered collection of all measurements for an entity. \\
Exemplar & $\Exemplar{x}$ & Self-contained synthetic dataset encoding a specific behavior \\
Global Dataset & $\GlobalDataset$ & The complete union of data chunks across the global time horizon. 
\end{tabular}
\end{footnotesize}
\centering
\vspace{-0.1cm}
\caption{Summary of  key terms and  notations}
\vspace{-0.3cm}

\label{tab:notation}
\end{table}
\subsection{Data Model and Preliminaries}
We begin with some  definitions and preliminaries to set the context and data model to ground the discussion. Table~\ref{tab:notation} summarizes the key concepts. 

\myparagraph{Entities and  Attributes} 
Let $\EntitySet = \{\EntityID{1}, \EntityID{2}, \ldots, \EntityID{n}\}$ be the finite set of
\emph{entities} in our world. For instance, these are users or user sessions in e-commerce analytics or sensors in IoT, or servers in  observability.
Each entity $\entity \in \EntitySet$ is characterized by a set of static attributes $\StaticAttrs = \{\staticKey_1, \staticKey_2, \ldots\}$. For instance,  the  OS version, location, or service type, or  user categories.  The value of attribute $\staticKey$ for entity $\entity$ is written as 
  $\staticVal{\entity}{\staticKey} \in \mathcal{D}_{\staticKey}$,
where $\mathcal{D}_{\staticKey}$ is the domain of key $\staticKey$.  The complete static
profile of entity $\entity$ is the tuple 
  $\StaticProfile{\entity} \;=\;
  \bigl(\staticVal{\entity}{\staticKey_1},\;
        \staticVal{\entity}{\staticKey_2},\;\ldots\bigr)$.

\myparagraph{Measurement Model}  A \emph{measurement} for entity $\entity$ at time $\ttime \in \TimeSet$ is a partial function
$\MeasPayload{\entity}{\ttime} \;:\; \DynKeys \rightharpoonup \bigcup_{\dynKey}\mathcal{D}_{\dynKey}$,
mapping a subset of dynamic attributes $\DynKeys = \{\dynKey_1, \dynKey_2, \ldots\}$ to
their observed values.  The measured value of a specific key $\dynKey$  is denoted
$
  \MeasKey{\entity}{\ttime}{\dynKey} \;\in\; \mathcal{D}_{\dynKey} \cup \{\bot\}$,
where $\bot$ indicates that no measurement was observed.  A single measurement is the set of key-value pairs
$\Measurement{\entity}{\ttime}
  \;=\;
  \bigl\{\,\KVPair{\dynKey}{v} \;\mid\;
    \dynKey \in \DynKeys,\;
    v = \MeasKey{\entity}{\ttime}{\dynKey} \ne \bot\,\bigr\}$.
The \emph{time series} of entity $\entity$ is the  ordered collection of 
timestamped measurements:
  $\TimeSeries{\entity}
  \;=\;
  \bigl\langle\,
    (\ttime,\,\Measurement{\entity}{\ttime})
    \;\mid\;
    \ttime \in \MeasSet{\entity}
  \,\bigr\rangle$,
where $\MeasSet{\entity} \subseteq \TimeSet$ is the (possibly aperiodic) set of observation
times for entity $\entity$.

\subsection{Data Generation}

\begin{figure}[ht]
    \centering
    \includegraphics[width=0.9\linewidth]{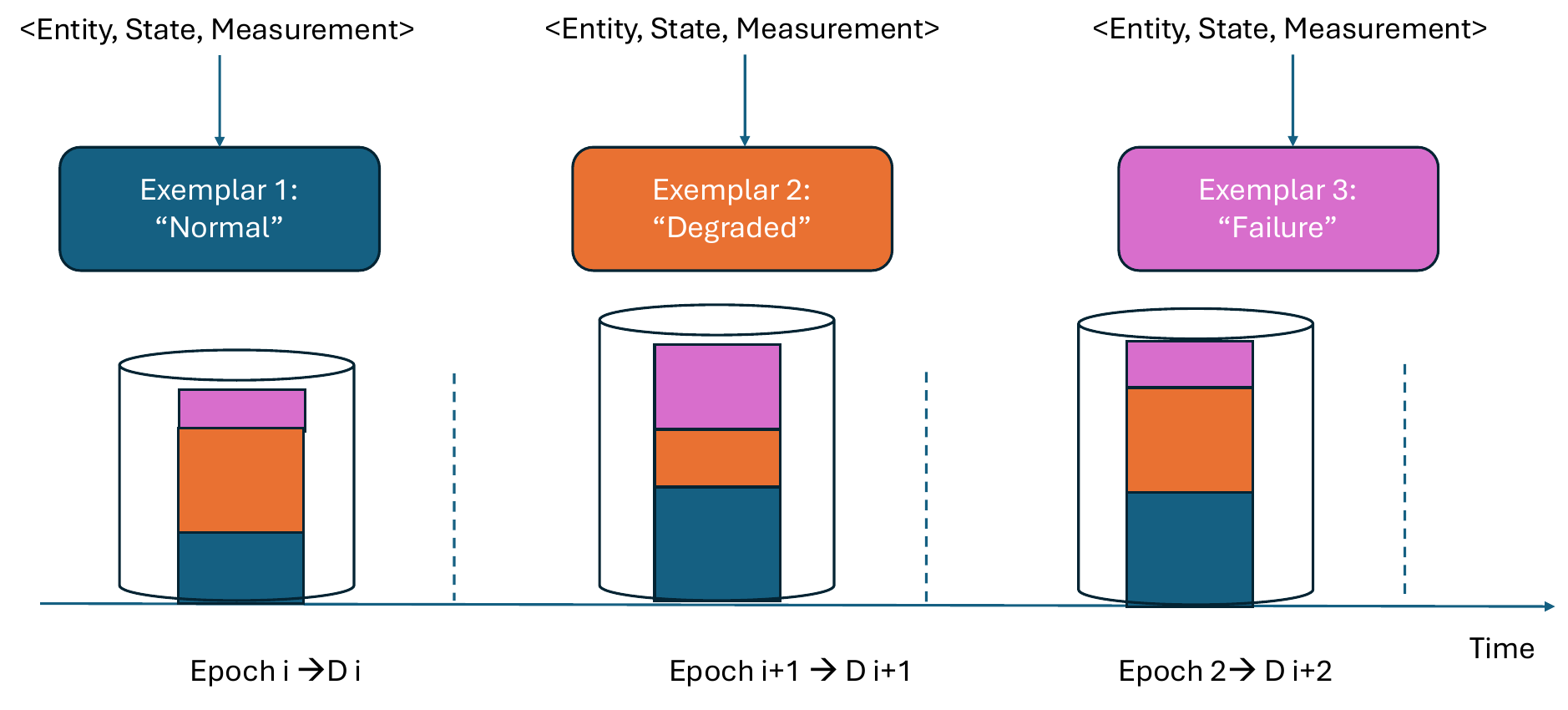}
       \vspace{-0.3cm}
    \caption{A simplified logical   overview of the generation process:  Create exemplars, and then blend them to create a  global timeseries of interest }
       \vspace{-0.3cm}
    \label{fig:sec4:overview}
\end{figure}

With these preliminaries, we can now define a modular, composable strategy for generating synthetic time series 
data.  Figure~\ref{fig:sec4:overview} shows an overview of the  generation process  in three logical steps:

\begin{packedenumerate}
  \item \textbf{Entity specifications} describe distributions over static entity attributes and   state
        transition diagrams governing dynamic  behavior, and per-state measurement generation.
  \item \textbf{Exemplar datasets} specify self-contained synthetic datasets, each encoding a
        particular behavior of interest by fixing an  entity  specification.
  \item \textbf{Global assembly} creates  a large-scale dataset  by partitioning time
         into epochs  and blending selected exemplars within each epoch  to produce a desired
        global pattern.
\end{packedenumerate}

\myparagraph{Entity  Specifications}  For each entity type and
static attribute key $\staticKey \in \StaticAttrs$, we  assign  a distribution to draw from: $\AttrDist{\staticKey}{\EntityType}
  \;=\;
  \AttrDistSpec{\mathcal{F}_{\staticKey}}{\boldsymbol{\theta}_{\staticKey}^{\EntityType}}{\ValDomain{\staticKey}}$,
where $\mathcal{F}_{\staticKey}$ is the distributional family (e.g.\ Normal, Uniform), $\boldsymbol{\theta}_{\staticKey}^{\EntityType}$ are its parameters, and
$\ValDomain{\staticKey}$ is the  domain.   Entities are partitioned into types $\EntityType \in \EntityTypeSet$ based on attributes as: 
$  \AttrSpec{\EntityType}
  \;=\;
  \bigl\{\,\staticKey \mapsto \AttrDist{\staticKey}{\EntityType} \;\mid\; \staticKey \in \StaticAttrs\,\bigr\}$.
A concrete entity $\entity$ of type $\EntityType$ is instantiated by sampling
$\staticVal{\entity}{\staticKey} \sim \AttrDist{\staticKey}{\EntityType}$ independently for
each $\staticKey$.

Each entity type is governed by a \emph{state transition distribution}: 
$\STD{\EntityType} = (\StateSet, \InitState, \TransFunc, \Delta)$ where, $\StateSet = \{\StateID{1}, \ldots, \StateID{m}\}$ is a finite set of named states
        (e.g.,$\mathrm{normal}$, $\mathrm{degraded}$,
        $\mathrm{failure}$),  $\InitState \in \StateSet$ is the initial state,  $\TransFunc$ encodes transition probabilities: $\TransProb{\entity}{\StateID{i}}{\StateID{j}}$
        is the probability of entity $\entity$ transitioning from $\StateID{i}$ to $\StateID{j}$,
        potentially conditioned on $\StaticProfile{\entity}$, and  $\StateDuration{}{}$ is the distribution over dwell times  in state $\mystate$ for entities of type $\EntityType$.

At any time $\ttime$, $\ActiveState{\entity}{\ttime} \in \StateSet$ denotes the current state
of entity $\entity$.  The state trajectory $\{\ActiveState{\entity}{\ttime}\}_{\ttime \ge 0}$
is a semi-Markov process driven by $\STD{\EntityType}$. Transition probabilities may depend on static attributes to encode
attribute--behavior relationships.

For each dynamic key $\dynKey \in \DynKeys$ and state $\mystate \in \StateSet$, the
\emph{measurement generation specification} $\MeasSpec{\dynKey}{\mystate}$ defines how
observations are drawn.  This specification 
  includes a signal structure, trend,     seasonality, and noise specification following classical timeseries models~\cite{chatfield1978holt}.  
The full generated measurement for entity $\entity$ at time $\ttime$ in state
$\mystate = \ActiveState{\entity}{\ttime}$ is
\[
  \MeasGenFn{\entity}{\ttime}{\mystate}
  \;=\;
  \bigl\{\,\KVPair{\dynKey}{\MeasKey{\entity}{\ttime}{\dynKey}}
          \;\mid\; \dynKey \in \DynKeys\,\bigr\}\]

\myparagraph{Exemplar Datasets}
An \emph{exemplar} $\Exemplar{x}$ is a fully specified, self-contained synthetic dataset
encoding a named behavior of interest $\ExemplarBehavior{x}$.  Its specification consists of the  entity type $\EntityType_x$  and the measurement spec: 
 \[\bigl(\,\EntityType_x,\;
         \AttrSpec\;,
         \STD\;, 
         \MeasSpec\;,
         \ExemplarEntities{x},\;
         \ExemplarDuration{x}
  \,\bigr)\]
where $\ExemplarEntities{x}$ is the number of entities to generate and $\ExemplarDuration{x}$
is the duration of the synthetic time interval.

To generate $\Exemplar{x}$ we use the following steps:
\begin{enumerate}
  \item Sample $\ExemplarEntities{x}$ entities, drawing static profiles
        \[\StaticProfile{\entity} \sim \AttrSpec{\EntityType_x}\]
  \item For each entity, simulate a state trajectory from \\
  $\STD{\EntityType_x}$ over $[0, \ExemplarDuration{x}]$.
  \item At each observation time $\ttime$ (drawn from an inter-arrival distribution), emit
        $\MeasGenFn{\entity}{\ttime}{\ActiveState{\entity}{\ttime}}$.
\end{enumerate}

The resulting exemplar dataset is
  \[
    \Exemplar{x}
    \;=\;
    \bigl\{\,
      \bigl(\entity,\,\StaticProfile{\entity},\,\TimeSeries{\entity}\bigr)
      \;\mid\;
      \entity \in \EntitySet_x
    \,\bigr\}
  \]
We can have an extensible library of available exemplars  $\ExemplarSet$.

\begin{table*}[ht]
\begin{footnotesize}
\begin{tabularx}{\textwidth}{
    p{0.1\textwidth}
    p{0.4\textwidth}
    >{\raggedright\arraybackslash}X
    p{0.5\textwidth}
  }
\toprule
 \textbf{Name} & \textbf{Formal Expression} & \textbf{Description} \\
\midrule
Event count &
$\Count{\entity,\EventPred{},\QueryWindow} = \bigl|\{\ttime \in \MeasSet{\entity}\cap\QueryWindow : \EventPred{}(\MeasPayload{\entity}{\ttime})\}\bigr|$ &
Integer count of events matching $\EventPred{}$ in the window. \\[4pt]

Event rate &
$\WinRate{\EventType{}}{\entity}{\QueryWindow} = \dfrac{\Count{\entity,\EventPred{},\QueryWindow}}{\tend - \tstart}$ &
Occurrence rate of a given event type per unit time. \\[4pt]

Attribute mean &
$\WinMean{\dynKey}{\entity}{\QueryWindow} = \dfrac{1}{\WinCount{\entity}{\QueryWindow}} \displaystyle\sum_{\ttime \in \MeasSet{\entity}\cap\QueryWindow} \MeasKey{\entity}{\ttime}{\dynKey}$ &
Mean of dynamic attribute $\dynKey$ over all matching events in $\QueryWindow$. \\[4pt]

Attribute std.\ dev. &
$\WinStd{\dynKey}{\entity}{\QueryWindow} = \sqrt{\dfrac{1}{N}\displaystyle\sum_{\ttime}\bigl(\MeasKey{\entity}{\ttime}{\dynKey} - \WinMean{\dynKey}{\entity}{\QueryWindow}\bigr)^{2}}$ &
Standard deviation of $\dynKey$ in $\QueryWindow$; measures signal volatility. \\[4pt]

Attribute percentile &
$\Percentile{p}{\dynKey,\entity,\QueryWindow}$ &
$p$-th percentile of $\dynKey$ values within the window (e.g.\ $p \in \{50,90,95\}$). \\[4pt]

Conditional aggregate &
$\Agg{OP}{\MeasKey{\entity}{\ttime}{\dynKey} : \ttime\in\QueryWindow,\,\EventPred{}(\MeasPayload{\entity}{\ttime})}$ &
Applies $\mathtt{OP} \in \{\mathrm{SUM},\mathrm{MAX},\mathrm{MIN},\mathrm{AVG}\}$ over events gated by predicate $\EventPred{}$. \\

\bottomrule
\end{tabularx}
\end{footnotesize}
\caption{Illustrative examples of stateless query templates. Here 
  $\entity$ refers to an entity; $\dynKey$ refers to a  dynamic attribute or measurement variable of interest, $\QueryWindow$ specifies a time  window, 
  $\EventPred{}$ is an event predicate, and $\EventType{A},\EventType{B}$ denote event types.}
\label{tab:stateless}
\end{table*}

\begin{table*}[ht]
\begin{footnotesize}
\begin{tabularx}{\textwidth}{
    p{0.1\textwidth}
    p{0.4\textwidth}
    >{\raggedright\arraybackslash}X
    p{0.5\textwidth}
  }
\toprule
\textbf{Name} & \textbf{Formal Expression} & \textbf{Description} \\
\midrule

Avg.\ time between events &
$\AvgTimeBetween{\EventType{A}}{\EventType{B}} = \dfrac{1}{|\mathcal{P}|}\displaystyle\sum_{(\ttime_A,\ttime_B)\in\mathcal{P}}(\ttime_B - \ttime_A)$ &
Mean latency between a trigger event $\EventType{A}$ and a subsequent event $\EventType{B}$, over matched pairs $\mathcal{P}$ within $\QueryWindow$. \\[4pt]

Sequence match &
$\SeqMatch{\EventSeq{\EventType{1},\ldots,\EventType{n}}}(\entity,\QueryWindow)$ &
Boolean: did the ordered sequence $\EventType{1}\prec\cdots\prec\EventType{n}$ occur in $\QueryWindow$, with optional inter-step time constraints? \\[4pt]

Count after trigger &
$\Count{\entity,\EventPred{\mathrm{tgt}},\QueryWindow \mid \Exists{\entity,\EventPred{\mathrm{trig}},\QueryWindow}}$ &
Count of target events occurring after a trigger event within $\QueryWindow$. \\[4pt]

Conversion rate &
$\ConvRate{\EventType{A}}{\EventType{B}}(\EntitySet',\QueryWindow) = \dfrac{|\{\entity\in\EntitySet' : \SeqMatch{\EventType{A},\EventType{B}}\}|}{|\{\entity : \Exists{\entity,\EventType{A},\QueryWindow}\}|}$ &
Fraction of entities that progressed from event $\EventType{A}$ to $\EventType{B}$; funnel conversion metric. \\[4pt]

Cross-window comparison &
$\CompareWindows{\dynKey}{\QueryWindow_1}{\QueryWindow_2} = \WinMean{\dynKey}{\entity}{\QueryWindow_2} - \WinMean{\dynKey}{\entity}{\QueryWindow_1}$ &
Signed change in a KPI between two windows; detects temporal drift. \\[4pt]

Alternating pattern count &
$\AltCount{\EventType{A},\EventType{B},\entity,\QueryWindow}$ &
Number of $\EventType{A}\!\to\!\EventType{B}\!\to\!\EventType{A}$ oscillation cycles in the window; detects vacillating behavior. \\[4pt]

\bottomrule
\end{tabularx}
\end{footnotesize}
\caption{Illustrative examples of stateful query templates that depend on the sequence and timing by reconstructing state machine semantics of the entity's behavior.}
\label{tab:stateful}
\end{table*}

We can create a number of pre-specified domain-specific $\ExemplarSet$ of interest.  For instance, for numerical timeseries in IoT/telecom, we can introduce patterns like {\em spikes}, {\em dips}, {\em slow growth}, or {\em data gaps} in the measurement. For product analytics, we can create $\ExemplarSet$ of various kinds of 
 user- or host activities; e.g., normal behavior of Windows hosts,  premium users purchasing a lot, hosts who abandon carts after adding products, and so on.  

\myparagraph{Global Dataset} We partition a global time horizon  into  ordered 
non-overlapping \emph{epochs} as shown in Figure~\ref{fig:overview}.
For each   $\Chunk{j}$, we create a  \emph{blend profile}
$  \BlendProfile{j}
  \;=\;
  \bigl(\BlendWeight{1}{j},\;\BlendWeight{2}{j},\;\ldots,\;\BlendWeight{|\ExemplarSet|}{j}\bigr),
  \qquad
  \BlendWeight{x}{j} \ge 0,\quad \sum_{x}\BlendWeight{x}{j} = 1$
specifying the relative contribution of each exemplar $\Exemplar{x}$. Given the  blend profile $\BlendProfile{j}$, the  dataset  $\ChunkDataset{j}$  is produced by
the blending function
\[ \BlendFn{j}{\BlendProfile{j}}
  \;=\;
  \bigsqcup_{x \;:\; \BlendWeight{x}{j} > 0}
  \;\mathrm{Sample}\!\bigl(\Exemplar{x},\;\lfloor\BlendWeight{x}{j}\cdot N_j\rfloor,\;\ChunkWindow{j}\bigr)\],
where $N_j$ is the total number of entity-time-series to include in chunk $\Chunk{j}$, and
$\mathrm{Sample}(\Exemplar{x}, n, I)$ draws $n$ entity trajectories from $\Exemplar{x}$,
time-shifted to interval $I$.

The global dataset is simply  the partitioned union across all chunks:
$  \GlobalDataset
  \;=\;
  \bigsqcup_{j=1}^{J} \ChunkDataset{j}$.
By varying $\BlendProfile{j}$ across epochs, we  can  encode a desired \emph{global
pattern} --- for example, a gradual drift from normal to degraded behavior, a seasonal
oscillation between exemplar types, or a sudden regime change.

  \subsection{Query-Answer Generation}

Given a generated $\GlobalDataset$, next we  focus on   generating {\em data aligned question-answer pairs} of interest to the domain. 
At a high-level, a {\em query} over the generated dataset $\mathbf{D}$ is specified as a triple 
\[
\QueryTemplate{} \;=\; \bigl(\EntityFilter{\cdot},\;\EventPredSet,\;\mathbf{F}\bigr),
\]
where $\EntityFilter{\cdot}$ is an optional predicate over static profiles
$\StaticProfile{\entity}$ that selects a sub-population of entities,
$\EventPredSet = \{\EventPred{1}, \ldots, \EventPred{m}\}$ is a set of event
predicates that filter or label individual measurements, and $\mathbf{F}$ is an
analysis function that aggregates the matched events into a result
$\QueryResult{}$.

Queries are evaluated against the windowed time series \linebreak
$\TimeSeriesWindow{\entity}{\tstart}{\tend}$ of each entity $\entity \in \EntitySet$.
The window $\QueryWindow$ may be either a  global time interval
$\TimeWin{\entity}{\tstart}{\tend}$ (e.g., Fri 9-10 am ET) or a predicated window
$\PredWin{\entity}{\ttime_0}{j}$ (e.g., when CPU measurement is high).  Results may
optionally be grouped  over static attribute values
$\GroupBy{a_1,\ldots,a_r}{\cdot}$.

We distinguish two broad families  of query templates:

\begin{packeditemize} 
  \item \textbf{Stateless trajectory-agnostic queries} treat the event
        stream inside $\QueryWindow$ as an {\em unordered collection}  of key-value payloads.
        They compute aggregate statistics — counts, rates, means, percentiles,
        inter-event durations — directly from $\MeasKey{\entity}{\ttime}{\dynKey}$. 
Stateless templates require only $\TimeSeriesWindow{\entity}{\tstart}{\tend}$
and the static profile $\StaticProfile{\entity}$.  They are evaluated
independently for each entity and window, with no shared state across
time steps. In other words, the outputs of the analysis do not depend on the sequence or timing of the events and view the data of each entity in a ``tabular'' fashion with rows being independent of each other. 
Table~\ref{tab:stateless} lists a few illustrative examples of stateless queries. For instance, we can ask if a particular event exists (e.g., did the user buy), or how many events of a type occurred (e.g., how many product  views), or a time-average  value (e.g., average CPU).

  \item \textbf{Stateful trajectory-dependent queries}   depend on the sequence and timing of events~\cite{cep}. At a high-level, they  require logically 
        replaying the event stream through a  logical state machine  
        $\STD{\EntityType}$, and    
        different analysis functions then
        operate on this state machine    to compute reachability,  durations,
        transition frequencies, common trajectory paths, and multi-state time
        distributions. 
     
        More specifically, these queries  first reconstruct  a \emph{state occupancy record}
for entity $\entity$ as 
\[
  \StateOcc{\entity}{\mystate}{\QueryWindow}
  \;=\;
  \bigl\{[\ttime^{\mathrm{in}}_j,\,\ttime^{\mathrm{out}}_j)
         \;\mid\;
         \ActiveState{\entity}{\ttime}=\mystate,\;
         \ttime\in\QueryWindow\bigr\},
\]
the set of contiguous intervals during which $\entity$ occupies $\mystate$ within
the window.  Entry and exit are governed by predicates $\StateEntry{\mystate}$ and
$\StateExit{\mystate}$, with optional timeout $\StateTimeout{\mystate}$. 
Table~\ref{tab:stateful} shows  illustrative  stateful queries. For instance, we calculate the time between events (e.g., between clicking on product to buying)   or check if a sequence of events occurred, or count how many events occurred in a state (e.g., how many outbound requests after compromise). Appendix~\ref{sec:appendix:morequeries} covers more query templates.

\end{packeditemize}

Given these basic building blocks of stateful and stateless query templates we can construct a rich library of question-answer pairs of interest. Note that the two families compose naturally: stateless KPIs may be computed
\emph{conditioned} on a state (e.g.\ mean sensor reading while in
$\StateID{critical}$), and stateful metrics may be further filtered by static
entity attributes.

\myparagraph{Incident-specific customization} 
These basic building blocks can be used to capture {\em incident-specific} queries of interest. An incident-specific query can be constructed by defining the three components of $\QueryTemplate{}$ as follows: the $\EntityFilter{}$ selects the affected sub-population of entities using predicates on their static profiles $\StaticProfile{}$, $\EventPredSet$ identifies the dynamic keys of interest; and the query window is set to a specific incident time interval $\TimeWin{}{}{}$. The analysis function $\mathbf{F}$ is applied to the matched measurements. It can be any of the previously described functions, or taken from a set of incident-related operations, e.g., incident existence checks, affected entity counts and rankings, and so on. To generate queries that require an agent to accurately find the incident in the data, we can systematically vary each component: selecting a different entity sub-population, a non-degraded dynamic key outside $\EventPredSet$, or shift the time interval to a baseline period disjoint from the incident’s time interval. Finally, incident queries can span multiple affected entities by chaining $\EntityFilter{}$ predicates, e.g., conditioning a query for $\entity_1$ on whether $\entity_2$ exhibits anomalous behavior.

\myparagraph{Language and persona variations} 
The agent developer may want to simulate different user personas (e.g., data engineer vs.\ SRE vs.\ VP vs.\  executive)  or different   dialects (e.g., customers in different geographical regions). 
 Once we have an initial question set and the associated  structured analysis (e.g., SQL or  code), we use a  LLM-assisted workflow to generate natural language variations of the questions. We use a simple LLM-as-a-judge approach to ensure that the variants  match the semantic intents of the question based on the reference question and analysis code.%

  \subsection{Test Integration}
 We envision \sysname being used in conjunction with existing test-driven development, and CI/CD  workflows  that  developers use today (e.g.,~\cite{opik, UK_AI_Security_Institute_Inspect_AI_Framework_2024}).  Integrating \sysname entails   two steps.  First, the dataset generated by \sysname needs to be made available the agent's backend; e.g., QA teams have a non-production test database or shadow tables. 
 Second, we need to issue the questions to the agent, receive the responses, and grade it. For the latter, we can either use a basic API response or provide the eval question-answer dataset to   frameworks~\cite{opik} or add it as  a benchmark on the data agent platform (e.g.,~\cite{geniebench}).

\section{Implementation}
\myparagraph{Dataset generation} We implement \sysname  data generation module (Phase 1 in Figure~\ref{fig:overview}) as a deterministic pipeline using a Python SDK that produces realistic domain-specific timeseries data. Entity behavior is driven by state transition diagrams and per-state measurements are generated via specifications.  
Self-contained exemplars encoding specific behaviors  are assembled into the global dataset $D_{\text{global}}$ via epoch-based blending as described.   For incident-specific customization, we develop an extensible and controllable pattern injection library that introduces common enterprise patterns (e.g., KPI degradation, data outages, sudden flash crowds).  To reduce effort, we use an LLM assistant  to  infer domain-specific schemas and entity specifications from the expert's input and to translate high-level intents into specifications.

\myparagraph{Question-Answer generation} We implement the question-answer generation  (Phase 2 in Figure~\ref{fig:overview}) to produce data-aligned question-answer pairs. Following the query model, each structured query  is executed deterministically against the dataset to produce ground truth results, covering both stateless  and stateful  queries.
For incident-specific queries, we construct structured  queries coupled to the injected patterns, e.g., queries that check for an incident or analyze  KPIs in the duration around it. We produce natural language questions  from the structured queries via a template-based  system, with optional LLM-assisted generation of  variations.

\myparagraph{Evaluation harness and setup} \sysname's  test integration exports the generated eval datasets and Q-A pairs for use in existing agent evaluation workflows. The curated dataset can be exported to database tables or flat files. The Q-A suite can be exported in structured formats compatible with external  tools and  development platforms. For our evaluation, we use a simple request-response harness that issues each question to the agent in a one-shot setting and records the natural language response for grading.

\myparagraph{Artifact and Reproducibility} The reference datasets, Q-A pairs, and agent benchmarks used here are available here: \url{https://huggingface.co/datasets/RockfishData/TimeSeriesAgentEvals}. Since \sysname is a proprietary tool, we can provide academic non-commercial access   for artifact evaluation. %

\section{Evaluation }
\label{sec:evaluation}

We demonstrate the value of \sysname by designing experiments that answer the following questions: (i) Do domain-specific benchmarks created using \sysname reveal gaps in timeseries data analysis agents that general benchmarks would miss? (ii) Why do SOTA agents fail on stateful and incident-specific queries? and (iii) Can we use \sysname to improve agent accuracy on these queries?

\subsection{Setup}

\myparagraph{Agents} We consider two proprietary agents (Databricks Genie, Snowflake Cortex Analyst) and two open-source agents (PandasAI, Nao) as representative examples of state-of-the-art data analysis agents. Since \sysname is agnostic to the agent implementation, i.e., the prompts, tools, and/or underlying LLMs, we do black-box testing by providing the agent access to the datasets and observing query-answer interactions in the one-shot setting.

For the Databricks and Snowflake agents, we used the default LLM that was available at account setup time \cite{dbrx2026genie,snowflakeanalyst}. We empirically observe that the Snowflake agent was using Sonnet 3.5. For the Nao agent, we used \texttt{gpt-4.1}. For PandasAI agents, we varied the underlying LLM to be one of: \texttt{o4-mini-2025-\allowbreak04-16}, \texttt{claude-sonnet-4-6}, and \texttt{claude-opus-4-6}. 

We provided the agents with dataset-specific context, such as the schema and data previews, using the recommended defaults in the setup process. We acknowledge that proprietary agents include additional features that enhance dataset-specific context, such as domain-specific metadata, example queries, and fine-tuning \cite{dbrx2026genie,snowflakeanalyst}. We defer experiments that show the effects of these features on agent performance to future work.

\begin{table}[th]
\begin{footnotesize}
\begin{tabular}{p{1.2cm}|p{1.7cm}|p{1cm}|p{1cm}|p{1.7cm}}
\textbf{Dataset} & \textbf{Description} & \textbf{\#Tables } & \textbf{\#Rows} & \textbf{Query Set} \\ \hline
E-commerce & User browsing sessions in an e-commerce setting & 2 & 6K & stateless, stateful  \\ 
IoT & IoT device health metrics for temperature, pressure, and humidity sensors & 1 & 50K & stateless, stateful \\ 
Telecom & Network telemetry for transport links, core nodes, and cell sites & 3 & 23.5K & stateless, stateful, incident \\ 
\end{tabular}
\end{footnotesize}
     \vspace{-0.1cm}
\caption{Overview of the three domain-specific \sysname test suites used for evaluation, including dataset dimensions and query types.}
     \vspace{-0.3cm}
\label{tab:datasets}
\end{table}

\myparagraph{Datasets and Query Sets} We use the same three  settings as in Section~\ref{sec:motivation}: product analytics for an e-commerce website (e-commerce), telecommunications network telemetry (telecom), and IoT device monitoring (IoT). For each setting, we create the domain-specific benchmark (dataset and query set) using \sysname.

First, we use the data generation module in \sysname to create datasets with the desired entities, attributes, and measurements. For the e-commerce dataset, we generate browsing sessions using a state machine that mainly generates browsing flows, along with a few cart abandonment and purchase flows. We create the IoT dataset using a sensor entity with three exemplars - temperature, pressure, and humidity sensors - that have their own operations state machines and device health metrics.  The telecom dataset consists of three entities related to one another: transport links, core nodes, and cell sites.  Each entity has its own attributes (e.g., location), and measurements (e.g. latency, availability). We additionally use the pattern injection library in \sysname to simulate a cascading incident in the telecom dataset: a transport link degrades (elevated packet loss, latency, jitter), cascading to connected cell sites (higher RRC failures, lower availability), and causing a modest effect on core nodes (reduced attached UEs, increased CPU load). 

Once the underlying datasets have been created, we use the question-answer generation module in \sysname to obtain the domain-specific query sets. The e-commerce and IoT query sets each contain 12 stateless queries and 12 stateful queries. The telecom query set contains 12 stateless and 12 stateless or stateful incident-specific queries. Table \ref{tab:datasets} provides a summary of the \sysname benchmarks used in our experiments.

\myparagraph{Methodology and Metrics} We record agent responses on every benchmark across 3 independent trials to counter the effects of non-deterministic agent behavior. Since agents often respond in natural language or return artifacts such as tables and charts,  for the accuracy evaluation we currently  manually code all agent responses. Each response is compared with the expected answer and is classified as one of three categories: correct answer, incorrect answer, or runtime error. The three answers for the same query are also assigned category labels. We report three metrics: \textbf{accuracy}, \textbf{\texttt{pass@2}}, and \textbf{self-consistency}. An agent response is accurate if its natural language response contains the expected answer, or if the expected answer can be derived from the artifacts. An agent response is considered inaccurate if it returns an incorrect answer or raises an error. Self-consistency is computed as the ratio of the majority category label count to the number of trials.

 \subsection{Key Findings}

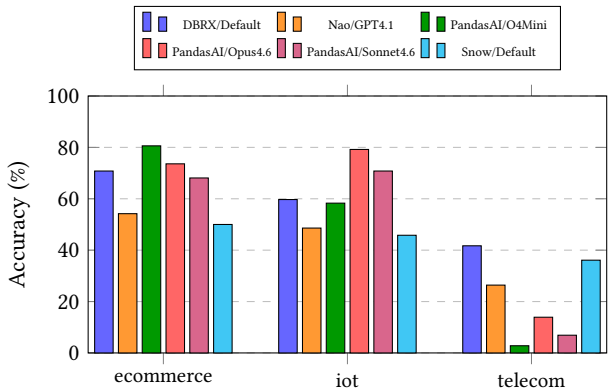
\begin{figure}[t]
  \centering
  \begin{tikzpicture}
\begin{axis}[
  ybar,
  bar width=7pt,
  width=\columnwidth,
  height=5cm,
  symbolic x coords={{ecommerce}, {iot}, {telecom}},
  xtick=data,
  xticklabel style={rotate=0, anchor=north, font=\small},
  enlarge x limits={abs=1cm},
  clip=false,
  ymin=0, ymax=100,
  ylabel={Accuracy (\%)},
  legend style={at={(0.5, 1.1)}, anchor=south, font=\tiny, legend columns=3},
  ymajorgrids=true,
  grid style=dashed,
]

  \addplot[fill=blue!60] coordinates {
    ({ecommerce}, 70.8)
    ({iot}, 59.7)
    ({telecom}, 41.7)
  };
  \addlegendentry{DBRX/Default}

  \addplot[fill=orange!80] coordinates {
    ({ecommerce}, 54.2)
    ({iot}, 48.6)
    ({telecom}, 26.4)
  };
  \addlegendentry{Nao/GPT4.1}

  \addplot[fill=green!60!black] coordinates {
    ({ecommerce}, 80.6)
    ({iot}, 58.3)
    ({telecom}, 2.8)
  };
  \addlegendentry{PandasAI/O4Mini}

  \addplot[fill=red!60] coordinates {
    ({ecommerce}, 73.6)
    ({iot}, 79.2)
    ({telecom}, 13.9)
  };
  \addlegendentry{PandasAI/Opus4.6}

  \addplot[fill=purple!60] coordinates {
    ({ecommerce}, 68.1)
    ({iot}, 70.8)
    ({telecom}, 6.9)
  };
  \addlegendentry{PandasAI/Sonnet4.6}

  \addplot[fill=cyan!60] coordinates {
    ({ecommerce}, 50.0)
    ({iot}, 45.8)
    ({telecom}, 36.1)
  };
  \addlegendentry{Snow/Default}

\end{axis}
\end{tikzpicture}
     \vspace{-0.3cm}
  \caption{Accuracy averaged over 3 runs, per dataset across agent/model combinations.}
     \vspace{-0.3cm}
  \label{fig:fig1_accuracy_dataset}
\end{figure}

\myparagraph{Overall Performance} Figure~\ref{fig:fig1_accuracy_dataset} shows the accuracy across agents and datasets. Overall, agents have good accuracy on e-commerce (66\%) and IoT (60\%) benchmarks, but struggle on the telecom (21\%) benchmark. The \texttt{pass@2} metric (Figure~\ref{fig:fig1_pass_2_dataset}, Appendix), which measures whether an agent gets the query right at least once when given two chances, does not show a substantial improvement. Combined with the self-consistency result (Figure ~\ref{fig:fig1_consistency_dataset}, Appendix), which shows that agents return the same response across trials 76\% of the time on average, this suggests that agents are not failing randomly; they have consistent  failure modes.

\begin{figure*}[htp]
  \centering
  \resizebox{0.8\textwidth}{!}{%
\begin{tikzpicture}
\begin{axis}[
  enlargelimits=false,
  colormap={metric}{color(0cm)=(red!70!white) color(50cm)=(yellow!80) color(100cm)=(green!60!black)},
  colorbar,
  colorbar style={
    ylabel={Accuracy (\%)},
    ytick={0,25,50,75,100},
  },
  point meta min=0,
  point meta max=100,
  width=14cm,
  height=5cm,
  xtick=\empty,
  ytick={0,1,2,3,4,5},
  yticklabels={DBRX/Default,Nao/GPT4.1,PandasAI/O4Mini,PandasAI/Opus4.6,PandasAI/Sonnet4.6,Snow/Default},
  yticklabel style={font=\small, anchor=east, text width=3cm, align=right},
  tick align=outside,
  axis on top,
  enlarge y limits={abs=0.5},
  clip=false
]
\addplot[matrix plot*, point meta=explicit, mesh/cols=24]
  coordinates {
  (0, 0) [100.0]
  (1, 0) [100.0]
  (2, 0) [100.0]
  (3, 0) [100.0]
  (4, 0) [100.0]
  (5, 0) [100.0]
  (6, 0) [100.0]
  (7, 0) [100.0]
  (8, 0) [100.0]
  (9, 0) [100.0]
  (10, 0) [100.0]
  (11, 0) [100.0]
  (12, 0) [100.0]
  (13, 0) [0.0]
  (14, 0) [0.0]
  (15, 0) [100.0]
  (16, 0) [33.3]
  (17, 0) [100.0]
  (18, 0) [100.0]
  (19, 0) [0.0]
  (20, 0) [66.7]
  (21, 0) [0.0]
  (22, 0) [0.0]
  (23, 0) [0.0]
  (0, 1) [100.0]
  (1, 1) [66.7]
  (2, 1) [66.7]
  (3, 1) [100.0]
  (4, 1) [100.0]
  (5, 1) [33.3]
  (6, 1) [66.7]
  (7, 1) [100.0]
  (8, 1) [100.0]
  (9, 1) [100.0]
  (10, 1) [100.0]
  (11, 1) [100.0]
  (12, 1) [0.0]
  (13, 1) [100.0]
  (14, 1) [0.0]
  (15, 1) [66.7]
  (16, 1) [33.3]
  (17, 1) [0.0]
  (18, 1) [33.3]
  (19, 1) [0.0]
  (20, 1) [33.3]
  (21, 1) [0.0]
  (22, 1) [0.0]
  (23, 1) [0.0]
  (0, 2) [100.0]
  (1, 2) [100.0]
  (2, 2) [100.0]
  (3, 2) [100.0]
  (4, 2) [100.0]
  (5, 2) [100.0]
  (6, 2) [100.0]
  (7, 2) [100.0]
  (8, 2) [100.0]
  (9, 2) [100.0]
  (10, 2) [100.0]
  (11, 2) [100.0]
  (12, 2) [100.0]
  (13, 2) [100.0]
  (14, 2) [0.0]
  (15, 2) [100.0]
  (16, 2) [0.0]
  (17, 2) [33.3]
  (18, 2) [100.0]
  (19, 2) [66.7]
  (20, 2) [100.0]
  (21, 2) [100.0]
  (22, 2) [33.3]
  (23, 2) [0.0]
  (0, 3) [100.0]
  (1, 3) [100.0]
  (2, 3) [100.0]
  (3, 3) [100.0]
  (4, 3) [100.0]
  (5, 3) [100.0]
  (6, 3) [100.0]
  (7, 3) [100.0]
  (8, 3) [100.0]
  (9, 3) [100.0]
  (10, 3) [100.0]
  (11, 3) [100.0]
  (12, 3) [33.3]
  (13, 3) [100.0]
  (14, 3) [100.0]
  (15, 3) [100.0]
  (16, 3) [33.3]
  (17, 3) [0.0]
  (18, 3) [0.0]
  (19, 3) [0.0]
  (20, 3) [66.7]
  (21, 3) [100.0]
  (22, 3) [0.0]
  (23, 3) [33.3]
  (0, 4) [100.0]
  (1, 4) [100.0]
  (2, 4) [100.0]
  (3, 4) [100.0]
  (4, 4) [100.0]
  (5, 4) [100.0]
  (6, 4) [100.0]
  (7, 4) [100.0]
  (8, 4) [100.0]
  (9, 4) [100.0]
  (10, 4) [100.0]
  (11, 4) [100.0]
  (12, 4) [100.0]
  (13, 4) [100.0]
  (14, 4) [66.7]
  (15, 4) [0.0]
  (16, 4) [0.0]
  (17, 4) [0.0]
  (18, 4) [0.0]
  (19, 4) [0.0]
  (20, 4) [66.7]
  (21, 4) [100.0]
  (22, 4) [0.0]
  (23, 4) [0.0]
  (0, 5) [100.0]
  (1, 5) [0.0]
  (2, 5) [100.0]
  (3, 5) [100.0]
  (4, 5) [100.0]
  (5, 5) [100.0]
  (6, 5) [100.0]
  (7, 5) [100.0]
  (8, 5) [100.0]
  (9, 5) [100.0]
  (10, 5) [100.0]
  (11, 5) [100.0]
  (12, 5) [0.0]
  (13, 5) [0.0]
  (14, 5) [0.0]
  (15, 5) [0.0]
  (16, 5) [0.0]
  (17, 5) [0.0]
  (18, 5) [0.0]
  (19, 5) [0.0]
  (20, 5) [100.0]
  (21, 5) [0.0]
  (22, 5) [0.0]
  (23, 5) [0.0]
  };

  \draw[white, line width=1.5pt] (axis cs:11.5,-0.5) -- (axis cs:11.5,5.5);

  \node[below=6pt, font=\small, rotate=0, anchor=north] at (axis cs:5.5,-0.5) {basic};
  \node[below=6pt, font=\small, rotate=0, anchor=north] at (axis cs:17.5,-0.5) {stateful};

\end{axis}
\end{tikzpicture}
}
     \vspace{-0.3cm}
  \caption{Per-query accuracy heatmap for the product/e-commerce dataset.}
     \vspace{-0.3cm}
  \label{fig:fig2_accuracy_ecommerce}
\end{figure*}
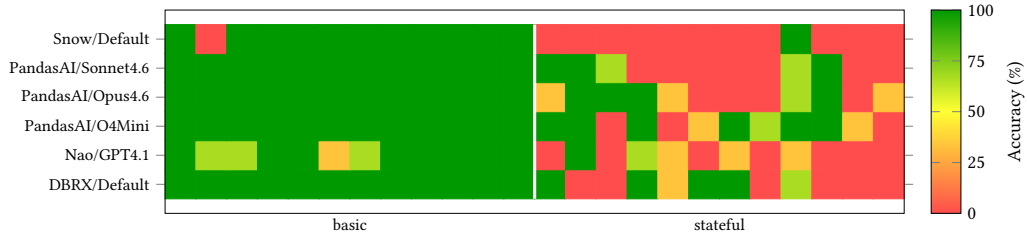

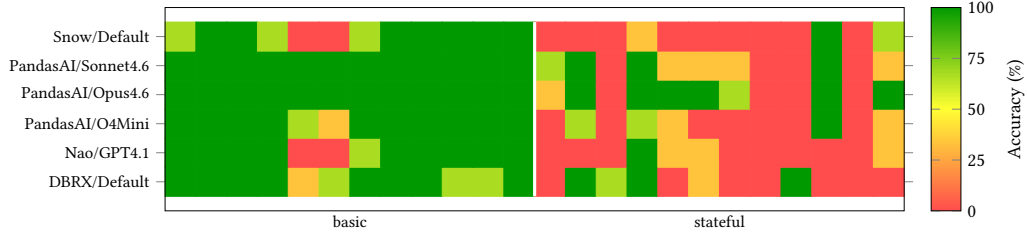
\begin{figure*}[htp]
  \centering
  \resizebox{0.8\textwidth}{!}{%
\begin{tikzpicture}
\begin{axis}[
  enlargelimits=false,
  colormap={metric}{color(0cm)=(red!70!white) color(50cm)=(yellow!80) color(100cm)=(green!60!black)},
  colorbar,
  colorbar style={
    ylabel={Accuracy (\%)},
    ytick={0,25,50,75,100},
  },
  point meta min=0,
  point meta max=100,
  width=14cm,
  height=5cm,
  xtick=\empty,
  ytick={0,1,2,3,4,5},
  yticklabels={DBRX/Default,Nao/GPT4.1,PandasAI/O4Mini,PandasAI/Opus4.6,PandasAI/Sonnet4.6,Snow/Default},
  yticklabel style={font=\small, anchor=east, text width=3cm, align=right},
  tick align=outside,
  axis on top,
  enlarge y limits={abs=0.5},
  clip=false
]
\addplot[matrix plot*, point meta=explicit, mesh/cols=24]
  coordinates {
  (0, 0) [100.0]
  (1, 0) [100.0]
  (2, 0) [100.0]
  (3, 0) [100.0]
  (4, 0) [33.3]
  (5, 0) [66.7]
  (6, 0) [100.0]
  (7, 0) [100.0]
  (8, 0) [100.0]
  (9, 0) [66.7]
  (10, 0) [66.7]
  (11, 0) [100.0]
  (12, 0) [0.0]
  (13, 0) [100.0]
  (14, 0) [66.7]
  (15, 0) [100.0]
  (16, 0) [0.0]
  (17, 0) [33.3]
  (18, 0) [0.0]
  (19, 0) [0.0]
  (20, 0) [100.0]
  (21, 0) [0.0]
  (22, 0) [0.0]
  (23, 0) [0.0]
  (0, 1) [100.0]
  (1, 1) [100.0]
  (2, 1) [100.0]
  (3, 1) [100.0]
  (4, 1) [0.0]
  (5, 1) [0.0]
  (6, 1) [66.7]
  (7, 1) [100.0]
  (8, 1) [100.0]
  (9, 1) [100.0]
  (10, 1) [100.0]
  (11, 1) [100.0]
  (12, 1) [0.0]
  (13, 1) [0.0]
  (14, 1) [0.0]
  (15, 1) [100.0]
  (16, 1) [33.3]
  (17, 1) [33.3]
  (18, 1) [0.0]
  (19, 1) [0.0]
  (20, 1) [0.0]
  (21, 1) [0.0]
  (22, 1) [0.0]
  (23, 1) [33.3]
  (0, 2) [100.0]
  (1, 2) [100.0]
  (2, 2) [100.0]
  (3, 2) [100.0]
  (4, 2) [66.7]
  (5, 2) [33.3]
  (6, 2) [100.0]
  (7, 2) [100.0]
  (8, 2) [100.0]
  (9, 2) [100.0]
  (10, 2) [100.0]
  (11, 2) [100.0]
  (12, 2) [0.0]
  (13, 2) [66.7]
  (14, 2) [0.0]
  (15, 2) [66.7]
  (16, 2) [33.3]
  (17, 2) [0.0]
  (18, 2) [0.0]
  (19, 2) [0.0]
  (20, 2) [0.0]
  (21, 2) [100.0]
  (22, 2) [0.0]
  (23, 2) [33.3]
  (0, 3) [100.0]
  (1, 3) [100.0]
  (2, 3) [100.0]
  (3, 3) [100.0]
  (4, 3) [100.0]
  (5, 3) [100.0]
  (6, 3) [100.0]
  (7, 3) [100.0]
  (8, 3) [100.0]
  (9, 3) [100.0]
  (10, 3) [100.0]
  (11, 3) [100.0]
  (12, 3) [33.3]
  (13, 3) [100.0]
  (14, 3) [0.0]
  (15, 3) [100.0]
  (16, 3) [100.0]
  (17, 3) [100.0]
  (18, 3) [66.7]
  (19, 3) [0.0]
  (20, 3) [0.0]
  (21, 3) [100.0]
  (22, 3) [0.0]
  (23, 3) [100.0]
  (0, 4) [100.0]
  (1, 4) [100.0]
  (2, 4) [100.0]
  (3, 4) [100.0]
  (4, 4) [100.0]
  (5, 4) [100.0]
  (6, 4) [100.0]
  (7, 4) [100.0]
  (8, 4) [100.0]
  (9, 4) [100.0]
  (10, 4) [100.0]
  (11, 4) [100.0]
  (12, 4) [66.7]
  (13, 4) [100.0]
  (14, 4) [0.0]
  (15, 4) [100.0]
  (16, 4) [33.3]
  (17, 4) [33.3]
  (18, 4) [33.3]
  (19, 4) [0.0]
  (20, 4) [0.0]
  (21, 4) [100.0]
  (22, 4) [0.0]
  (23, 4) [33.3]
  (0, 5) [66.7]
  (1, 5) [100.0]
  (2, 5) [100.0]
  (3, 5) [66.7]
  (4, 5) [0.0]
  (5, 5) [0.0]
  (6, 5) [66.7]
  (7, 5) [100.0]
  (8, 5) [100.0]
  (9, 5) [100.0]
  (10, 5) [100.0]
  (11, 5) [100.0]
  (12, 5) [0.0]
  (13, 5) [0.0]
  (14, 5) [0.0]
  (15, 5) [33.3]
  (16, 5) [0.0]
  (17, 5) [0.0]
  (18, 5) [0.0]
  (19, 5) [0.0]
  (20, 5) [0.0]
  (21, 5) [100.0]
  (22, 5) [0.0]
  (23, 5) [66.7]
  };

  \draw[white, line width=1.5pt] (axis cs:11.5,-0.5) -- (axis cs:11.5,5.5);

  \node[below=6pt, font=\small, rotate=0, anchor=north] at (axis cs:5.5,-0.5) {basic};
  \node[below=6pt, font=\small, rotate=0, anchor=north] at (axis cs:17.5,-0.5) {stateful};

\end{axis}
\end{tikzpicture}
}
     \vspace{-0.3cm}
  \caption{Per-query accuracy heatmap for the IoT dataset.}
     \vspace{-0.3cm}
  \label{fig:fig2_accuracy_iot}
\end{figure*}

\begin{figure*}[htp]
  \centering
  \resizebox{0.8\textwidth}{!}{%
\begin{tikzpicture}
\begin{axis}[
  enlargelimits=false,
  colormap={metric}{color(0cm)=(red!70!white) color(50cm)=(yellow!80) color(100cm)=(green!60!black)},
  colorbar,
  colorbar style={
    ylabel={Accuracy (\%)},
    ytick={0,25,50,75,100},
  },
  point meta min=0,
  point meta max=100,
  width=14cm,
  height=5cm,
  xtick=\empty,
  ytick={0,1,2,3,4,5},
  yticklabels={DBRX/Default,Nao/GPT4.1,PandasAI/O4Mini,PandasAI/Opus4.6,PandasAI/Sonnet4.6,Snow/Default},
  yticklabel style={font=\small, anchor=east, text width=3cm, align=right},
  tick align=outside,
  axis on top,
  enlarge y limits={abs=0.5},
  clip=false
]
\addplot[matrix plot*, point meta=explicit, mesh/cols=24]
  coordinates {
  (0, 0) [100.0]
  (1, 0) [66.7]
  (2, 0) [0.0]
  (3, 0) [66.7]
  (4, 0) [100.0]
  (5, 0) [100.0]
  (6, 0) [100.0]
  (7, 0) [66.7]
  (8, 0) [100.0]
  (9, 0) [33.3]
  (10, 0) [66.7]
  (11, 0) [100.0]
  (12, 0) [100.0]
  (13, 0) [0.0]
  (14, 0) [0.0]
  (15, 0) [0.0]
  (16, 0) [0.0]
  (17, 0) [0.0]
  (18, 0) [0.0]
  (19, 0) [0.0]
  (20, 0) [0.0]
  (21, 0) [0.0]
  (22, 0) [0.0]
  (23, 0) [0.0]
  (0, 1) [100.0]
  (1, 1) [0.0]
  (2, 1) [33.3]
  (3, 1) [66.7]
  (4, 1) [33.3]
  (5, 1) [33.3]
  (6, 1) [66.7]
  (7, 1) [100.0]
  (8, 1) [66.7]
  (9, 1) [0.0]
  (10, 1) [0.0]
  (11, 1) [100.0]
  (12, 1) [33.3]
  (13, 1) [0.0]
  (14, 1) [0.0]
  (15, 1) [0.0]
  (16, 1) [0.0]
  (17, 1) [0.0]
  (18, 1) [0.0]
  (19, 1) [0.0]
  (20, 1) [0.0]
  (21, 1) [0.0]
  (22, 1) [0.0]
  (23, 1) [0.0]
  (0, 2) [0.0]
  (1, 2) [0.0]
  (2, 2) [0.0]
  (3, 2) [0.0]
  (4, 2) [0.0]
  (5, 2) [0.0]
  (6, 2) [0.0]
  (7, 2) [0.0]
  (8, 2) [0.0]
  (9, 2) [0.0]
  (10, 2) [0.0]
  (11, 2) [0.0]
  (12, 2) [33.3]
  (13, 2) [0.0]
  (14, 2) [0.0]
  (15, 2) [0.0]
  (16, 2) [0.0]
  (17, 2) [0.0]
  (18, 2) [33.3]
  (19, 2) [0.0]
  (20, 2) [0.0]
  (21, 2) [0.0]
  (22, 2) [0.0]
  (23, 2) [0.0]
  (0, 3) [0.0]
  (1, 3) [0.0]
  (2, 3) [0.0]
  (3, 3) [0.0]
  (4, 3) [0.0]
  (5, 3) [0.0]
  (6, 3) [0.0]
  (7, 3) [0.0]
  (8, 3) [0.0]
  (9, 3) [0.0]
  (10, 3) [0.0]
  (11, 3) [0.0]
  (12, 3) [33.3]
  (13, 3) [0.0]
  (14, 3) [0.0]
  (15, 3) [0.0]
  (16, 3) [33.3]
  (17, 3) [100.0]
  (18, 3) [33.3]
  (19, 3) [0.0]
  (20, 3) [0.0]
  (21, 3) [66.7]
  (22, 3) [33.3]
  (23, 3) [33.3]
  (0, 4) [0.0]
  (1, 4) [0.0]
  (2, 4) [0.0]
  (3, 4) [0.0]
  (4, 4) [0.0]
  (5, 4) [0.0]
  (6, 4) [0.0]
  (7, 4) [0.0]
  (8, 4) [0.0]
  (9, 4) [0.0]
  (10, 4) [0.0]
  (11, 4) [0.0]
  (12, 4) [100.0]
  (13, 4) [33.3]
  (14, 4) [0.0]
  (15, 4) [0.0]
  (16, 4) [0.0]
  (17, 4) [0.0]
  (18, 4) [0.0]
  (19, 4) [0.0]
  (20, 4) [0.0]
  (21, 4) [0.0]
  (22, 4) [33.3]
  (23, 4) [0.0]
  (0, 5) [100.0]
  (1, 5) [0.0]
  (2, 5) [0.0]
  (3, 5) [100.0]
  (4, 5) [100.0]
  (5, 5) [100.0]
  (6, 5) [100.0]
  (7, 5) [100.0]
  (8, 5) [66.7]
  (9, 5) [0.0]
  (10, 5) [100.0]
  (11, 5) [100.0]
  (12, 5) [0.0]
  (13, 5) [0.0]
  (14, 5) [0.0]
  (15, 5) [0.0]
  (16, 5) [0.0]
  (17, 5) [0.0]
  (18, 5) [0.0]
  (19, 5) [0.0]
  (20, 5) [0.0]
  (21, 5) [0.0]
  (22, 5) [0.0]
  (23, 5) [0.0]
  };

  \draw[white, line width=1.5pt] (axis cs:11.5,-0.5) -- (axis cs:11.5,5.5);

  \node[below=6pt, font=\small, rotate=0, anchor=north] at (axis cs:5.5,-0.5) {basic};
  \node[below=6pt, font=\small, rotate=0, anchor=north] at (axis cs:17.5,-0.5) {incident};

\end{axis}
\end{tikzpicture}
}
     \vspace{-0.3cm}
  \caption{Per-query accuracy heatmap for the telecom dataset.}
     \vspace{-0.3cm}
  \label{fig:fig2_accuracy_telecom}
\end{figure*}
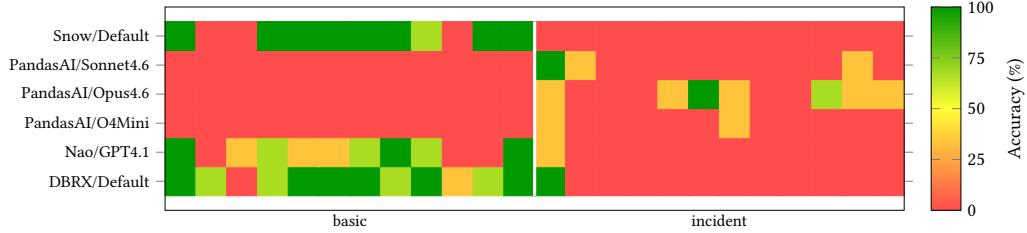

\myparagraph{Per-Query Type Performance} We next  analyze whether agents perform differently across query types. Figures~\ref{fig:fig2_accuracy_ecommerce} -- ~\ref{fig:fig2_accuracy_telecom} show the break down accuracy by query type for the three datasets. Agents consistently perform well on stateless queries (73\%), but performance drops significantly on stateful queries (34\%), and further still on incident-specific queries (10\%). The effect is most pronounced in the telecom benchmark, where several agents get nearly no stateful queries correct. These results confirm our anecdotal findings from Section~\ref{sec:motivation}: domain-specific, stateful queries expose a capability gap that general benchmarks do not. 

\myparagraph{Differences Across Agents and Datasets} We find that the PandasAI agents perform the best, which we attribute to two reasons. The first reason is the PandasAI agent can write both Python and SQL code to analyze data. Greater expressivity compared to traditional SQL might be giving it a performance boost, especially on stateful queries~\cite{milner2023raising}. The second reason could be that the underlying LLMs used by the PandasAI agents (Sonnet 4.6, Opus 4.6, and O4Mini) are newer than those used by the other agents. For example, our Snowflake agent performs worst (44\%), possibly due to its older underlying model (Sonnet 3.5). Surprisingly, among the PandasAI agents, we find that the more ``capable’’ LLMs (e.g. Opus 4.6) perform worse on our benchmarks than ``simpler’’ ones (e.g., O4Mini), showing how the current methods of measuring overall model progress might be hiding granular failures \cite{rabanser2026towards}. Across the datasets, the telecom benchmark is the hardest for two possible reasons: (i) it spans multiple tables, requiring agents to understand which combination(s) of tables need to be analyzed, and (ii) its query set contains incident-specific questions, requiring agents to understand which slices of the data constitute anomalous behavior.

\subsection{Failure Case Studies}

We now investigate representative failure cases to characterize the error modes across query types. We look for queries that most agents got incorrect from the following categories: stateless, stateful (no incident), and incident. The primary failure mode for stateless queries is schema confusion: agents either infer the wrong time interval (Listings~\ref{lst:dbrx_telecom_basic_error} and~\ref{lst:snow_telecom_basic_error}), or, as with all PandasAI variants, select the wrong table entirely and report that the queried column does not exist (Listing~\ref{lst:pandasai_basic_error}). 

Stateful (no incident) queries expose a different weakness: agents struggle to track state across events within a session. On an ecommerce benchmark query that asks for the number of product views while a cart is full, most agents issue a SQL query that does not update the view count while keeping track of the preceding add to cart events, leading to inflated counts (Listings~\ref{lst:dbrx_ecommerce_stateful_no_inc_error} and~\ref{lst:snow_ecommerce_stateful_no_inc_error}). 

For incident-related queries (both stateless and stateful), the dominant failure mode is the absence of data exploration: agents assume a fixed or global incident time window rather than detecting the incident from the data, and none attempt to pinpoint the affected entities (Listings~\ref{lst:nao_telecom_stateless_inc_error} 
and~\ref{lst:dbrx_telecom_stateless_inc_error}). Intuitively, an incident should be detected by comparing an entity against its own historical behavior, and not against a global average (Listing~\ref{lst:pandasai_sonnet_telecom_stateless_inc_error}). However, all agents apply thresholds or compare with baseline slices that are not derived from the data (Listings~\ref{lst:nao_telecom_stateful_inc_error}, 
\ref{lst:pandasai_opus_telecom_stateful_inc_error}, 
and~\ref{lst:dbrx_telecom_stateful_inc_error}). This suggests that augmenting timeseries data agents with specialized tools or operations to capture incident semantics could improve performance.

\subsection{Using \sysname to Improve Agents}

As a preliminary study, we evaluate how \sysname can be used in conjunction with off-the-shelf prompt optimization techniques like GEPA~\cite{agrawal2026gepareflectivepromptevolution}. 
We ran a preliminary  experiment on the telecom dataset using PandasAI agents, instantiated with the same three models as before: Sonnet 4.6, Opus 4.6, and O4Mini. The query set from Section~\ref{sec:evaluation} serves as the test set. We generate 30 additional queries using  \sysname for GEPA to use as its training and validation sets. We configure GEPA with a budget of 200 evaluations, specify  agent performance as the objective, and use \texttt{gpt-4.1} as the reflection model.\footnote{Compared to the main evaluation, we made two changes to ensure compatibility with GEPA: we constrained the agent's output to a single string response  and we implement an LLM-as-a-judge verifier using \texttt{gpt-4o-mini} to automate scoring.}.
 Overall, we  find that GEPA-optimized prompts using \sysname evals  improve performance by 17\% (Figure~\ref{fig:fig3_gepa_accuracy_improvement}, Appendix). Listing~\ref{lst:gepa_optim_prompt} shows the optimized prompt for the PandasAI O4Mini agent, which showed a 25\% accuracy improvement. This preliminary result suggests a further value of \sysname beyond testing to improving agent performance.

\section{Related Work} 

We describe related work in different aspects of data agents. At a high level, our focus is on timeseries agents in domains such as IoT, monitoring, telecommunications, and we find that there is a key gap in this respect.

Recent data agent benchmarks primarily focus on generic tabular tasks. Tapilot-Crossing \cite{tapilot2024} uses a multi-agent environment but is limited to Kaggle datasets lacking temporal complexity. DA-Code \cite{dacode2024}, InfiAgent-DABench \cite{infiagent2024}, and DABstep \cite{dabstep2025} span the data science lifecycle and offer verifiable closed-form questions, but lack domain-specific customization for observability or security contexts. ConDABench \cite{condabench2025} advances conversational evaluation through multi-turn interactions, but for a more generic problem domain. Reliability concerns noted by \cite{benchmarkingds2024} and \cite{jin2026cidr} highlight issues such as misinterpreted data types and the difficulty in establishing ground truth for incident-driven scenarios. Traditional benchmarks such as Spider remain foundational, yet Spider 2.0 \cite{lei2024spider2} and recent work on annotation errors \cite{jin2026pervasive, schmidt2025sqlstorm, liu2025nl2sqlbugs, luoma2025snails} suggest they may no longer capture the complexity of real-world enterprise workflows. The SQL2NL paper \cite{sql2nl2025} demonstrates that models are brittle to linguistic variations \cite{rahaman2024evaluating}, and general surveys \cite{agentsurvey2024, evalsurvey2025, luo2025large, rabanser2026towards, anthropic2026demystifying, pan2025measuring} identify a lack of specialized benchmarks for enterprise workflows and call for automated, scalable evaluation techniques. While \cite{liu2025empowering} provides an overview of how synthetic data is used to evaluate timeseries foundation models in \cite{cai2024timeseriesexam, merrill2024language, ashok2024context, aksu2024xforecast, gruver2023large}, the paper highlights the data is not realistic enough, i.e., it does not contain domain-specific patterns or anomalous behavior. Most existing efforts do not prescribe a methodology for creating custom evals for timeseries agents in new domains. Our work addresses this gap. 

Modern data agents rely on reasoning-action loops such as \textit{ReAct} \cite{yao2023react} and Reflexion \cite{shinn2023reflexion}, and are deployed in industry tools like Databricks Genie~\cite{databricksgenie}, Snowflake Cortex~\cite{snowflakeanalyst}, Grafana \cite{grafana-assistant}, and others \cite{hextech, openaiagent2024, uberquerygpt, chen2025enterprise, insightagents2026}. Evaluation frameworks such as IntellAgent \cite{intellagent2025} target conversational AI but do not address domain-specific timeseries analytics. While these frameworks enable agents to interact with databases via Text-to-SQL or Python code execution, they lack customizable evaluation for timeseries analytics specifically. Recent work on knowledge augmentation and agentic memory, including Mem0 \cite{chhikara2025mem0}, AgentSM \cite{biswal2026agentsm}, Knowledge Base Construction \cite{baek2025knowledge}, and MemOS \cite{li2025memos}, addresses schema understanding and long-term persistence for complex database environments \cite{xu2025mem, zhong2024memorybank, agarwal2026tribal}. \sysname complements these efforts by providing a methodology to generate domain-customized datasets, incidents, and query patterns to evaluate agent reliability.

\section{Conclusions and Future Work}

We view \sysname as a first but significant step in advancing  evals for timeseries analysis agents. We conclude by acknowledging  several limitations and directions for improvement.  First, we focus only on the {\em analysis}  workflow and ignore  other issues in data wrangling or data cleaning.  Second, we focus on {\em one-shot} agents that answer the  question  as posed; a natural direction  is  to extend \sysname to agents that can ask questions to refine the intent.  
Third, we showed the value of \sysname as a testing framework. As future work, we plan to  provide more fine-grained debugging (e.g., possibly connecting with backend tracing) to  understand  failure patterns and continuous training.

\bibliographystyle{abbrv}
\bibliography{agenteval}

\clearpage
\newpage

\appendix 
\section{Additional Evaluation Results}

\begin{figure}[htp]
  \centering
  \begin{tikzpicture}
\begin{axis}[
  ybar,
  bar width=7pt,
  width=\columnwidth,
  height=5cm,
  symbolic x coords={{ecommerce}, {iot}, {telecom}},
  xtick=data,
  xticklabel style={rotate=0, anchor=north, font=\small},
  enlarge x limits={abs=1cm},
  clip=false,
  ymin=0, ymax=100,
  ylabel={Pass@2 (\%)},
  legend style={at={(0.5, 1.1)}, anchor=south, font=\tiny, legend columns=3},
  ymajorgrids=true,
  grid style=dashed,
]

  \addplot[fill=blue!60] coordinates {
    ({ecommerce}, 73.6)
    ({iot}, 68.1)
    ({telecom}, 48.6)
  };
  \addlegendentry{DBRX/Default}

  \addplot[fill=orange!80] coordinates {
    ({ecommerce}, 65.3)
    ({iot}, 54.2)
    ({telecom}, 36.1)
  };
  \addlegendentry{Nao/GPT4.1}

  \addplot[fill=green!60!black] coordinates {
    ({ecommerce}, 84.7)
    ({iot}, 66.7)
    ({telecom}, 5.6)
  };
  \addlegendentry{PandasAI/O4Mini}

  \addplot[fill=red!60] coordinates {
    ({ecommerce}, 79.2)
    ({iot}, 81.9)
    ({telecom}, 22.2)
  };
  \addlegendentry{PandasAI/Opus4.6}

  \addplot[fill=purple!60] coordinates {
    ({ecommerce}, 70.8)
    ({iot}, 77.8)
    ({telecom}, 9.7)
  };
  \addlegendentry{PandasAI/Sonnet4.6}

  \addplot[fill=cyan!60] coordinates {
    ({ecommerce}, 50.0)
    ({iot}, 52.8)
    ({telecom}, 37.5)
  };
  \addlegendentry{Snow/Default}

\end{axis}
\end{tikzpicture}
  \caption{\texttt{pass@2} for 3 runs, per dataset across agent/model combinations.}
  \label{fig:fig1_pass_2_dataset}
\end{figure}
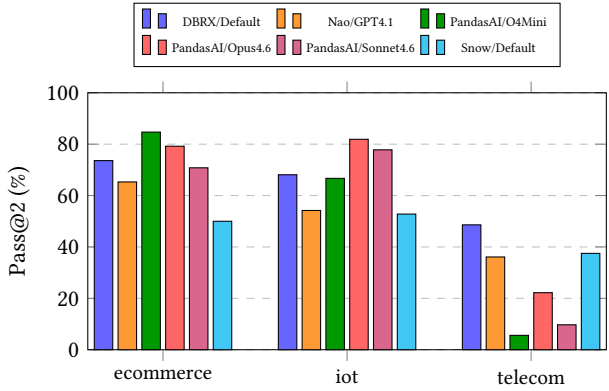

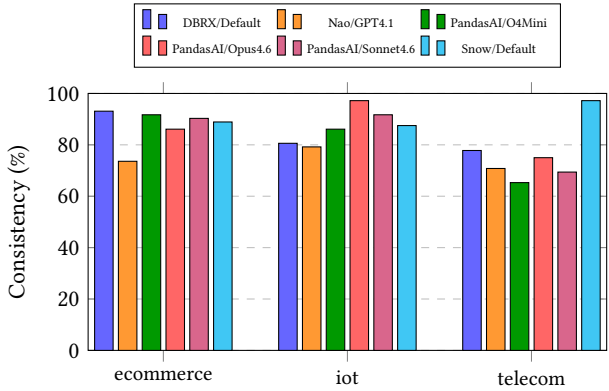
\begin{figure}[htp]
  \centering
  \begin{tikzpicture}
\begin{axis}[
  ybar,
  bar width=7pt,
  width=\columnwidth,
  height=5cm,
  symbolic x coords={{ecommerce}, {iot}, {telecom}},
  xtick=data,
  xticklabel style={rotate=0, anchor=north, font=\small},
  enlarge x limits={abs=1cm},
  clip=false,
  ymin=0, ymax=100,
  ylabel={Consistency (\%)},
  legend style={at={(0.5, 1.1)}, anchor=south, font=\tiny, legend columns=3},
  ymajorgrids=true,
  grid style=dashed,
]

  \addplot[fill=blue!60] coordinates {
    ({ecommerce}, 93.1)
    ({iot}, 80.6)
    ({telecom}, 77.8)
  };
  \addlegendentry{DBRX/Default}

  \addplot[fill=orange!80] coordinates {
    ({ecommerce}, 73.6)
    ({iot}, 79.2)
    ({telecom}, 70.8)
  };
  \addlegendentry{Nao/GPT4.1}

  \addplot[fill=green!60!black] coordinates {
    ({ecommerce}, 91.7)
    ({iot}, 86.1)
    ({telecom}, 65.3)
  };
  \addlegendentry{PandasAI/O4Mini}

  \addplot[fill=red!60] coordinates {
    ({ecommerce}, 86.1)
    ({iot}, 97.2)
    ({telecom}, 75.0)
  };
  \addlegendentry{PandasAI/Opus4.6}

  \addplot[fill=purple!60] coordinates {
    ({ecommerce}, 90.3)
    ({iot}, 91.7)
    ({telecom}, 69.4)
  };
  \addlegendentry{PandasAI/Sonnet4.6}

  \addplot[fill=cyan!60] coordinates {
    ({ecommerce}, 88.9)
    ({iot}, 87.5)
    ({telecom}, 97.2)
  };
  \addlegendentry{Snow/Default}

\end{axis}
\end{tikzpicture}
  \caption{Self-consistency per dataset across agent/model combinations.}
\label{fig:fig1_consistency_dataset}
\end{figure}

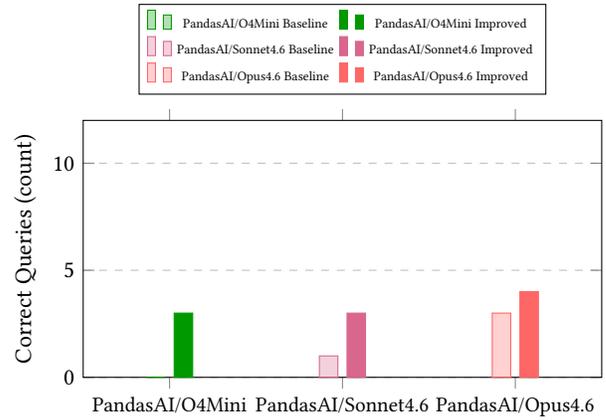
\begin{figure}[htp]
  \centering
  \begin{tikzpicture}
\begin{axis}[
  ybar,
  bar width=7pt,
  width=\columnwidth,
  height=5cm,
  xtick={1, 2, 3},
  xticklabels={{PandasAI/O4Mini}, {PandasAI/Sonnet4.6}, {PandasAI/Opus4.6}},
  xticklabel style={rotate=0, anchor=north, font=\small},
  xmin=0.5, xmax=3.5,
  clip=false,
  ymin=0, ymax=12,
  ylabel={Correct Queries (count)},
  legend style={at={(0.5, 1.1)}, anchor=south, font=\tiny, legend columns=2},
  ymajorgrids=true,
  grid style=dashed,
]

  \addplot[fill=green!60!black, fill opacity=0.3, draw=green!60!black, bar shift=0pt] coordinates {(0.92, 0)};
  \addlegendentry{PandasAI/O4Mini Baseline}

  \addplot[fill=green!60!black, fill opacity=1.0, draw=green!60!black, bar shift=0pt] coordinates {(1.08, 3)};
  \addlegendentry{PandasAI/O4Mini Improved}

  \addplot[fill=purple!60, fill opacity=0.3, draw=purple!60, bar shift=0pt] coordinates {(1.92, 1)};
  \addlegendentry{PandasAI/Sonnet4.6 Baseline}

  \addplot[fill=purple!60, fill opacity=1.0, draw=purple!60, bar shift=0pt] coordinates {(2.08, 3)};
  \addlegendentry{PandasAI/Sonnet4.6 Improved}

  \addplot[fill=red!60, fill opacity=0.3, draw=red!60, bar shift=0pt] coordinates {(2.92, 3)};
  \addlegendentry{PandasAI/Opus4.6 Baseline}

  \addplot[fill=red!60, fill opacity=1.0, draw=red!60, bar shift=0pt] coordinates {(3.08, 4)};
  \addlegendentry{PandasAI/Opus4.6 Improved}

\end{axis}
\end{tikzpicture}
  \caption{Improvement in telecom incident-specific queries across agents after using GEPA.}
  \label{fig:fig3_gepa_accuracy_improvement}
\end{figure}

\begin{listing}[htp]
\begin{minted}[breaklines, frame=single, linenos]{text}
You are a data analyst. Answer ONLY with a single number or a short text answer, whichever is most accurate. Do NOT include explanations, lists, tables, dicts, DataFrames, charts, or extra formatting. Read each question carefully and make sure your answer exactly matches the calculation, value, or label requested, with correct units and precision.
\end{minted}
\caption{Example of a GEPA-optimized prompt for a PandasAI O4Mini agent. This prompt showed a 25\% accuracy improvement on the telecom benchmark's incident-specific query set.}
\label{lst:gepa_optim_prompt}
\end{listing}

\clearpage

\newpage 

\onecolumn

\section{Detailed Results on AgentFuel Benchmarks}
\label{sec:appendix:morefail}

\begin{longtable}{>{\raggedright\arraybackslash}p{0.38\textwidth} p{1.7cm} c p{1.7cm} c p{1.7cm} c}
\toprule
\textbf{Query} & \multicolumn{2}{c}{\textbf{Run 1}} & \multicolumn{2}{c}{\textbf{Run 2}} & \multicolumn{2}{c}{\textbf{Run 3}} \\
\cmidrule(lr){2-3} \cmidrule(lr){4-5} \cmidrule(lr){6-7}
 & Class & Label & Class & Label & Class & Label \\
\midrule
\endfirsthead
\toprule
\textbf{Query} & \multicolumn{2}{c}{\textbf{Run 1}} & \multicolumn{2}{c}{\textbf{Run 2}} & \multicolumn{2}{c}{\textbf{Run 3}} \\
\cmidrule(lr){2-3} \cmidrule(lr){4-5} \cmidrule(lr){6-7}
 & Class & Label & Class & Label & Class & Label \\
\midrule
\endhead
\midrule
\multicolumn{7}{r}{\small\textit{(continued)}} \\
\endfoot
\bottomrule
\caption{ecommerce / DBRX/Default / stateful}
\endlastfoot
How many users added more than 4 items within 1 hour to the cart but exited without purchasing? & correct & 1 & correct & 1 & correct & 1 \\
Did premium users make purchases? & incorrect & 1 & incorrect & 1 & incorrect & 1 \\
What is the average time between viewing a product and adding it to cart? & incorrect & 1 & correct & 2 & incorrect & 1 \\
Did users view, add to cart, checkout, and then purchase? & correct & 1 & correct & 1 & correct & 1 \\
How many product views after cart abandonment? & correct & 1 & correct & 1 & correct & 1 \\
Did any users add to cart but not checkout within 5 minutes? & incorrect & 1 & incorrect & 1 & incorrect & 1 \\
Did any users view products more than 3 times without adding to cart? & correct & 1 & incorrect & 2 & correct & 1 \\
Which users reached the checkout stage, and where were these users acquired from? & incorrect & 1 & incorrect & 1 & incorrect & 1 \\
How many product views occurred while users had an item in their cart? & incorrect & 1 & incorrect & 2 & incorrect & 1 \\
How long on average do users spend with items in cart before purchasing or abandoning? & incorrect & 1 & incorrect & 2 & incorrect & 1 \\
What are the common state transitions from browsing to purchase or abandonment? & incorrect & 1 & incorrect & 2 & incorrect & 2 \\
What percentage of time do users spend in each page? & correct & 1 & correct & 1 & correct & 1 \\
\end{longtable}

\begin{longtable}{>{\raggedright\arraybackslash}p{0.38\textwidth} p{1.7cm} c p{1.7cm} c p{1.7cm} c}
\toprule
\textbf{Query} & \multicolumn{2}{c}{\textbf{Run 1}} & \multicolumn{2}{c}{\textbf{Run 2}} & \multicolumn{2}{c}{\textbf{Run 3}} \\
\cmidrule(lr){2-3} \cmidrule(lr){4-5} \cmidrule(lr){6-7}
 & Class & Label & Class & Label & Class & Label \\
\midrule
\endfirsthead
\toprule
\textbf{Query} & \multicolumn{2}{c}{\textbf{Run 1}} & \multicolumn{2}{c}{\textbf{Run 2}} & \multicolumn{2}{c}{\textbf{Run 3}} \\
\cmidrule(lr){2-3} \cmidrule(lr){4-5} \cmidrule(lr){6-7}
 & Class & Label & Class & Label & Class & Label \\
\midrule
\endhead
\midrule
\multicolumn{7}{r}{\small\textit{(continued)}} \\
\endfoot
\bottomrule
\caption{ecommerce / Snow/Default / stateful}
\endlastfoot
How many users added more than 4 items within 1 hour to the cart but exited without purchasing? & incorrect & 1 & incorrect & 1 & incorrect & 2 \\
Did premium users make purchases? & incorrect & 1 & incorrect & 2 & incorrect & 2 \\
What is the average time between viewing a product and adding it to cart? & incorrect & 1 & incorrect & 2 & incorrect & 1 \\
Did users view, add to cart, checkout, and then purchase? & incorrect & 1 & incorrect & 2 & incorrect & 2 \\
How many product views after cart abandonment? & incorrect & 1 & incorrect & 2 & incorrect & 1 \\
Did any users add to cart but not checkout within 5 minutes? & incorrect & 1 & incorrect & 2 & incorrect & 1 \\
Did any users view products more than 3 times without adding to cart? & correct & 1 & correct & 1 & correct & 1 \\
Which users reached the checkout stage, and where were these users acquired from? & incorrect & 1 & incorrect & 1 & incorrect & 1 \\
How many product views occurred while users had an item in their cart? & incorrect & 1 & incorrect & 1 & incorrect & 1 \\
How long on average do users spend with items in cart before purchasing or abandoning? & incorrect & 1 & incorrect & 2 & incorrect & 2 \\
What are the common state transitions from browsing to purchase or abandonment? & incorrect & 1 & incorrect & 1 & incorrect & 2 \\
What percentage of time do users spend in each page? & incorrect & 1 & incorrect & 1 & incorrect & 1 \\
\end{longtable}

\begin{longtable}{>{\raggedright\arraybackslash}p{0.38\textwidth} p{1.7cm} c p{1.7cm} c p{1.7cm} c}
\toprule
\textbf{Query} & \multicolumn{2}{c}{\textbf{Run 1}} & \multicolumn{2}{c}{\textbf{Run 2}} & \multicolumn{2}{c}{\textbf{Run 3}} \\
\cmidrule(lr){2-3} \cmidrule(lr){4-5} \cmidrule(lr){6-7}
 & Class & Label & Class & Label & Class & Label \\
\midrule
\endfirsthead
\toprule
\textbf{Query} & \multicolumn{2}{c}{\textbf{Run 1}} & \multicolumn{2}{c}{\textbf{Run 2}} & \multicolumn{2}{c}{\textbf{Run 3}} \\
\cmidrule(lr){2-3} \cmidrule(lr){4-5} \cmidrule(lr){6-7}
 & Class & Label & Class & Label & Class & Label \\
\midrule
\endhead
\midrule
\multicolumn{7}{r}{\small\textit{(continued)}} \\
\endfoot
\bottomrule
\caption{ecommerce / PandasAI/Opus4.6 / stateful}
\endlastfoot
How many users added more than 4 items within 1 hour to the cart but exited without purchasing? & incorrect & 1 & correct & 2 & incorrect & 3 \\
Did premium users make purchases? & correct & 1 & correct & 1 & correct & 1 \\
What is the average time between viewing a product and adding it to cart? & correct & 1 & incorrect & 2 & incorrect & 2 \\
Did users view, add to cart, checkout, and then purchase? & incorrect & 1 & incorrect & 2 & incorrect & 3 \\
How many product views after cart abandonment? & incorrect & 1 & incorrect & 2 & incorrect & 3 \\
Did any users add to cart but not checkout within 5 minutes? & incorrect & 1 & incorrect & 1 & incorrect & 1 \\
Did any users view products more than 3 times without adding to cart? & correct & 1 & incorrect & 2 & correct & 1 \\
Which users reached the checkout stage, and where were these users acquired from? & correct & 1 & correct & 1 & correct & 1 \\
How many product views occurred while users had an item in their cart? & incorrect & 1 & incorrect & 1 & incorrect & 1 \\
How long on average do users spend with items in cart before purchasing or abandoning? & incorrect & 1 & incorrect & 2 & correct & 3 \\
What are the common state transitions from browsing to purchase or abandonment? & correct & 1 & correct & 1 & correct & 1 \\
What percentage of time do users spend in each page? & correct & 1 & correct & 1 & correct & 1 \\
\end{longtable}

\begin{longtable}{>{\raggedright\arraybackslash}p{0.38\textwidth} p{1.7cm} c p{1.7cm} c p{1.7cm} c}
\toprule
\textbf{Query} & \multicolumn{2}{c}{\textbf{Run 1}} & \multicolumn{2}{c}{\textbf{Run 2}} & \multicolumn{2}{c}{\textbf{Run 3}} \\
\cmidrule(lr){2-3} \cmidrule(lr){4-5} \cmidrule(lr){6-7}
 & Class & Label & Class & Label & Class & Label \\
\midrule
\endfirsthead
\toprule
\textbf{Query} & \multicolumn{2}{c}{\textbf{Run 1}} & \multicolumn{2}{c}{\textbf{Run 2}} & \multicolumn{2}{c}{\textbf{Run 3}} \\
\cmidrule(lr){2-3} \cmidrule(lr){4-5} \cmidrule(lr){6-7}
 & Class & Label & Class & Label & Class & Label \\
\midrule
\endhead
\midrule
\multicolumn{7}{r}{\small\textit{(continued)}} \\
\endfoot
\bottomrule
\caption{ecommerce / Nao/GPT4.1 / stateful}
\endlastfoot
How many users added more than 4 items within 1 hour to the cart but exited without purchasing? & incorrect & 1 & incorrect & 2 & incorrect & 3 \\
Did premium users make purchases? & correct & 1 & correct & 1 & correct & 1 \\
What is the average time between viewing a product and adding it to cart? & incorrect & 1 & runtime\_error & 2 & correct & 3 \\
Did users view, add to cart, checkout, and then purchase? & runtime\_error & 1 & runtime\_error & 1 & incorrect & 2 \\
How many product views after cart abandonment? & incorrect & 1 & correct & 2 & incorrect & 3 \\
Did any users add to cart but not checkout within 5 minutes? & runtime\_error & 1 & incorrect & 2 & incorrect & 2 \\
Did any users view products more than 3 times without adding to cart? & incorrect & 1 & incorrect & 1 & correct & 2 \\
Which users reached the checkout stage, and where were these users acquired from? & runtime\_error & 1 & incorrect & 2 & runtime\_error & 1 \\
How many product views occurred while users had an item in their cart? & incorrect & 1 & incorrect & 1 & incorrect & 1 \\
How long on average do users spend with items in cart before purchasing or abandoning? & runtime\_error & 1 & incorrect & 2 & incorrect & 3 \\
What are the common state transitions from browsing to purchase or abandonment? & incorrect & 1 & incorrect & 2 & runtime\_error & 3 \\
What percentage of time do users spend in each page? & correct & 1 & correct & 1 & incorrect & 2 \\
\end{longtable}

\begin{longtable}{>{\raggedright\arraybackslash}p{0.38\textwidth} p{1.7cm} c p{1.7cm} c p{1.7cm} c}
\toprule
\textbf{Query} & \multicolumn{2}{c}{\textbf{Run 1}} & \multicolumn{2}{c}{\textbf{Run 2}} & \multicolumn{2}{c}{\textbf{Run 3}} \\
\cmidrule(lr){2-3} \cmidrule(lr){4-5} \cmidrule(lr){6-7}
 & Class & Label & Class & Label & Class & Label \\
\midrule
\endfirsthead
\toprule
\textbf{Query} & \multicolumn{2}{c}{\textbf{Run 1}} & \multicolumn{2}{c}{\textbf{Run 2}} & \multicolumn{2}{c}{\textbf{Run 3}} \\
\cmidrule(lr){2-3} \cmidrule(lr){4-5} \cmidrule(lr){6-7}
 & Class & Label & Class & Label & Class & Label \\
\midrule
\endhead
\midrule
\multicolumn{7}{r}{\small\textit{(continued)}} \\
\endfoot
\bottomrule
\caption{iot / DBRX/Default / stateful}
\endlastfoot
How many sensors exceeded a warning threshold more than 3 times in 12 hours without triggering maintenance? & incorrect & 1 & incorrect & 1 & incorrect & 2 \\
Did sensors on firmware v1.0 ever reach critical state? & correct & 1 & correct & 1 & correct & 1 \\
What is the average time between threshold exceeded and maintenance required? & incorrect & 1 & incorrect & 2 & incorrect & 3 \\
Were any v2.0 sensors in warning, critical, maintenance, and then operational? & incorrect & 1 & incorrect & 2 & correct & 3 \\
How many readings were recorded after maintenance was required? & incorrect & 1 & incorrect & 1 & incorrect & 1 \\
Did any sensors go from warning to critical without maintenance within an hour? & incorrect & 1 & incorrect & 1 & incorrect & 1 \\
Did any v1.1 sensors raise warnings more than 3 times without ever reaching a critical status? & correct & 1 & correct & 1 & correct & 1 \\
Which sensors reached critical status, and what location zone were they in? & incorrect & 1 & incorrect & 1 & incorrect & 1 \\
How many readings were recorded while sensors were being maintained? Show me a breakdown by device type. & incorrect & 1 & incorrect & 2 & incorrect & 3 \\
How long on average do devices stay in critical status before entering maintenance? Show me a breakdown by device type. & incorrect & 1 & incorrect & 2 & incorrect & 3 \\
What are the most common state transitions from degraded status to operational or offline? Show me the results according to the device location. & correct & 1 & correct & 1 & incorrect & 2 \\
What percentage of time do devices spend in each status? Show me a breakdown by device type. & correct & 1 & correct & 1 & correct & 1 \\
\end{longtable}

\begin{longtable}{>{\raggedright\arraybackslash}p{0.38\textwidth} p{1.7cm} c p{1.7cm} c p{1.7cm} c}
\toprule
\textbf{Query} & \multicolumn{2}{c}{\textbf{Run 1}} & \multicolumn{2}{c}{\textbf{Run 2}} & \multicolumn{2}{c}{\textbf{Run 3}} \\
\cmidrule(lr){2-3} \cmidrule(lr){4-5} \cmidrule(lr){6-7}
 & Class & Label & Class & Label & Class & Label \\
\midrule
\endfirsthead
\toprule
\textbf{Query} & \multicolumn{2}{c}{\textbf{Run 1}} & \multicolumn{2}{c}{\textbf{Run 2}} & \multicolumn{2}{c}{\textbf{Run 3}} \\
\cmidrule(lr){2-3} \cmidrule(lr){4-5} \cmidrule(lr){6-7}
 & Class & Label & Class & Label & Class & Label \\
\midrule
\endhead
\midrule
\multicolumn{7}{r}{\small\textit{(continued)}} \\
\endfoot
\bottomrule
\caption{iot / Snow/Default / stateful}
\endlastfoot
How many sensors exceeded a warning threshold more than 3 times in 12 hours without triggering maintenance? & incorrect & 1 & incorrect & 1 & incorrect & 1 \\
Did sensors on firmware v1.0 ever reach critical state? & incorrect & 1 & incorrect & 1 & incorrect & 1 \\
What is the average time between threshold exceeded and maintenance required? & incorrect & 1 & incorrect & 2 & incorrect & 2 \\
Were any v2.0 sensors in warning, critical, maintenance, and then operational? & incorrect & 1 & incorrect & 1 & incorrect & 2 \\
How many readings were recorded after maintenance was required? & incorrect & 1 & incorrect & 1 & incorrect & 1 \\
Did any sensors go from warning to critical without maintenance within an hour? & runtime\_error & 1 & runtime\_error & 1 & runtime\_error & 1 \\
Did any v1.1 sensors raise warnings more than 3 times without ever reaching a critical status? & incorrect & 1 & incorrect & 1 & incorrect & 1 \\
Which sensors reached critical status, and what location zone were they in? & correct & 1 & correct & 1 & correct & 1 \\
How many readings were recorded while sensors were being maintained? Show me a breakdown by device type. & incorrect & 1 & incorrect & 1 & incorrect & 1 \\
How long on average do devices stay in critical status before entering maintenance? Show me a breakdown by device type. & correct & 1 & correct & 1 & incorrect & 2 \\
What are the most common state transitions from degraded status to operational or offline? Show me the results according to the device location. & incorrect & 1 & incorrect & 1 & runtime\_error & 2 \\
What percentage of time do devices spend in each status? Show me a breakdown by device type. & correct & 1 & incorrect & 2 & runtime\_error & 3 \\
\end{longtable}

\begin{longtable}{>{\raggedright\arraybackslash}p{0.38\textwidth} p{1.7cm} c p{1.7cm} c p{1.7cm} c}
\toprule
\textbf{Query} & \multicolumn{2}{c}{\textbf{Run 1}} & \multicolumn{2}{c}{\textbf{Run 2}} & \multicolumn{2}{c}{\textbf{Run 3}} \\
\cmidrule(lr){2-3} \cmidrule(lr){4-5} \cmidrule(lr){6-7}
 & Class & Label & Class & Label & Class & Label \\
\midrule
\endfirsthead
\toprule
\textbf{Query} & \multicolumn{2}{c}{\textbf{Run 1}} & \multicolumn{2}{c}{\textbf{Run 2}} & \multicolumn{2}{c}{\textbf{Run 3}} \\
\cmidrule(lr){2-3} \cmidrule(lr){4-5} \cmidrule(lr){6-7}
 & Class & Label & Class & Label & Class & Label \\
\midrule
\endhead
\midrule
\multicolumn{7}{r}{\small\textit{(continued)}} \\
\endfoot
\bottomrule
\caption{iot / PandasAI/Opus4.6 / stateful}
\endlastfoot
How many sensors exceeded a warning threshold more than 3 times in 12 hours without triggering maintenance? & correct & 1 & incorrect & 2 & incorrect & 2 \\
Did sensors on firmware v1.0 ever reach critical state? & correct & 1 & correct & 1 & correct & 1 \\
What is the average time between threshold exceeded and maintenance required? & correct & 1 & correct & 1 & correct & 1 \\
Were any v2.0 sensors in warning, critical, maintenance, and then operational? & correct & 1 & correct & 1 & correct & 1 \\
How many readings were recorded after maintenance was required? & correct & 1 & incorrect & 2 & correct & 1 \\
Did any sensors go from warning to critical without maintenance within an hour? & runtime\_error & 1 & runtime\_error & 1 & runtime\_error & 1 \\
Did any v1.1 sensors raise warnings more than 3 times without ever reaching a critical status? & incorrect & 1 & incorrect & 1 & incorrect & 1 \\
Which sensors reached critical status, and what location zone were they in? & correct & 1 & correct & 1 & correct & 1 \\
How many readings were recorded while sensors were being maintained? Show me a breakdown by device type. & incorrect & 1 & incorrect & 1 & incorrect & 1 \\
How long on average do devices stay in critical status before entering maintenance? Show me a breakdown by device type. & correct & 1 & correct & 1 & correct & 1 \\
What are the most common state transitions from degraded status to operational or offline? Show me the results according to the device location. & incorrect & 1 & incorrect & 1 & incorrect & 1 \\
What percentage of time do devices spend in each status? Show me a breakdown by device type. & correct & 1 & correct & 1 & correct & 1 \\
\end{longtable}

\begin{longtable}{>{\raggedright\arraybackslash}p{0.38\textwidth} p{1.7cm} c p{1.7cm} c p{1.7cm} c}
\toprule
\textbf{Query} & \multicolumn{2}{c}{\textbf{Run 1}} & \multicolumn{2}{c}{\textbf{Run 2}} & \multicolumn{2}{c}{\textbf{Run 3}} \\
\cmidrule(lr){2-3} \cmidrule(lr){4-5} \cmidrule(lr){6-7}
 & Class & Label & Class & Label & Class & Label \\
\midrule
\endfirsthead
\toprule
\textbf{Query} & \multicolumn{2}{c}{\textbf{Run 1}} & \multicolumn{2}{c}{\textbf{Run 2}} & \multicolumn{2}{c}{\textbf{Run 3}} \\
\cmidrule(lr){2-3} \cmidrule(lr){4-5} \cmidrule(lr){6-7}
 & Class & Label & Class & Label & Class & Label \\
\midrule
\endhead
\midrule
\multicolumn{7}{r}{\small\textit{(continued)}} \\
\endfoot
\bottomrule
\caption{iot / Nao/GPT4.1 / stateful}
\endlastfoot
How many sensors exceeded a warning threshold more than 3 times in 12 hours without triggering maintenance? & runtime\_error & 1 & incorrect & 2 & incorrect & 2 \\
Did sensors on firmware v1.0 ever reach critical state? & incorrect & 1 & incorrect & 1 & incorrect & 1 \\
What is the average time between threshold exceeded and maintenance required? & correct & 1 & incorrect & 2 & incorrect & 3 \\
Were any v2.0 sensors in warning, critical, maintenance, and then operational? & incorrect & 1 & correct & 2 & runtime\_error & 3 \\
How many readings were recorded after maintenance was required? & incorrect & 1 & runtime\_error & 2 & incorrect & 3 \\
Did any sensors go from warning to critical without maintenance within an hour? & runtime\_error & 1 & incorrect & 2 & incorrect & 3 \\
Did any v1.1 sensors raise warnings more than 3 times without ever reaching a critical status? & incorrect & 1 & runtime\_error & 2 & incorrect & 1 \\
Which sensors reached critical status, and what location zone were they in? & incorrect & 1 & runtime\_error & 2 & incorrect & 1 \\
How many readings were recorded while sensors were being maintained? Show me a breakdown by device type. & incorrect & 1 & runtime\_error & 2 & runtime\_error & 2 \\
How long on average do devices stay in critical status before entering maintenance? Show me a breakdown by device type. & runtime\_error & 1 & incorrect & 2 & correct & 1 \\
What are the most common state transitions from degraded status to operational or offline? Show me the results according to the device location. & incorrect & 1 & incorrect & 1 & incorrect & 1 \\
What percentage of time do devices spend in each status? Show me a breakdown by device type. & correct & 1 & correct & 1 & correct & 1 \\
\end{longtable}

\begin{longtable}{>{\raggedright\arraybackslash}p{0.38\textwidth} p{1.7cm} c p{1.7cm} c p{1.7cm} c}
\toprule
\textbf{Query} & \multicolumn{2}{c}{\textbf{Run 1}} & \multicolumn{2}{c}{\textbf{Run 2}} & \multicolumn{2}{c}{\textbf{Run 3}} \\
\cmidrule(lr){2-3} \cmidrule(lr){4-5} \cmidrule(lr){6-7}
 & Class & Label & Class & Label & Class & Label \\
\midrule
\endfirsthead
\toprule
\textbf{Query} & \multicolumn{2}{c}{\textbf{Run 1}} & \multicolumn{2}{c}{\textbf{Run 2}} & \multicolumn{2}{c}{\textbf{Run 3}} \\
\cmidrule(lr){2-3} \cmidrule(lr){4-5} \cmidrule(lr){6-7}
 & Class & Label & Class & Label & Class & Label \\
\midrule
\endhead
\midrule
\multicolumn{7}{r}{\small\textit{(continued)}} \\
\endfoot
\bottomrule
\caption{telecom / DBRX/Default / incident}
\endlastfoot
Did the router that caused the outage have elevated packet loss on January 2 morning? & correct & 1 & correct & 1 & correct & 1 \\
What was the average latency on the affected router link during the incident? & incorrect & 1 & incorrect & 2 & incorrect & 1 \\
How many transport links were experiencing high packet loss during the outage? & incorrect & 1 & incorrect & 2 & incorrect & 2 \\
Which transport link had the worst latency during the outage? & incorrect & 1 & incorrect & 2 & incorrect & 1 \\
Were connection failures elevated on cells behind the affected router during the outage? & incorrect & 1 & incorrect & 2 & incorrect & 3 \\
What was the average connection failure rate on cells behind the affected router during the outage? & incorrect & 1 & incorrect & 2 & incorrect & 3 \\
How many cells lost availability during the outage on January 2? & incorrect & 1 & incorrect & 2 & incorrect & 2 \\
Which 5 cells had the most connection failures during the outage? & incorrect & 1 & incorrect & 1 & incorrect & 1 \\
Were connection failures elevated on cells behind the affected router during the outage, while the core nodes were also under load? & incorrect & 1 & incorrect & 2 & runtime\_error & 3 \\
Were connection failures elevated on cells behind the affected router during the outage, even though core node session counts looked normal? & incorrect & 1 & incorrect & 2 & incorrect & 2 \\
How many cells lost availability during the outage on January 2, while the core nodes were also under load? & incorrect & 1 & incorrect & 1 & incorrect & 1 \\
How many cells lost availability during the outage on January 2, compared to the same time the day before when core nodes were healthy? & incorrect & 1 & incorrect & 1 & incorrect & 1 \\
\end{longtable}

\begin{longtable}{>{\raggedright\arraybackslash}p{0.38\textwidth} p{1.7cm} c p{1.7cm} c p{1.7cm} c}
\toprule
\textbf{Query} & \multicolumn{2}{c}{\textbf{Run 1}} & \multicolumn{2}{c}{\textbf{Run 2}} & \multicolumn{2}{c}{\textbf{Run 3}} \\
\cmidrule(lr){2-3} \cmidrule(lr){4-5} \cmidrule(lr){6-7}
 & Class & Label & Class & Label & Class & Label \\
\midrule
\endfirsthead
\toprule
\textbf{Query} & \multicolumn{2}{c}{\textbf{Run 1}} & \multicolumn{2}{c}{\textbf{Run 2}} & \multicolumn{2}{c}{\textbf{Run 3}} \\
\cmidrule(lr){2-3} \cmidrule(lr){4-5} \cmidrule(lr){6-7}
 & Class & Label & Class & Label & Class & Label \\
\midrule
\endhead
\midrule
\multicolumn{7}{r}{\small\textit{(continued)}} \\
\endfoot
\bottomrule
\caption{telecom / Snow/Default / incident}
\endlastfoot
Did the router that caused the outage have elevated packet loss on January 2 morning? & incorrect & 1 & incorrect & 1 & incorrect & 1 \\
What was the average latency on the affected router link during the incident? & incorrect & 1 & incorrect & 1 & incorrect & 1 \\
How many transport links were experiencing high packet loss during the outage? & incorrect & 1 & incorrect & 1 & incorrect & 1 \\
Which transport link had the worst latency during the outage? & incorrect & 1 & incorrect & 1 & incorrect & 1 \\
Were connection failures elevated on cells behind the affected router during the outage? & incorrect & 1 & incorrect & 1 & incorrect & 1 \\
What was the average connection failure rate on cells behind the affected router during the outage? & incorrect & 1 & incorrect & 1 & incorrect & 2 \\
How many cells lost availability during the outage on January 2? & incorrect & 1 & incorrect & 1 & incorrect & 1 \\
Which 5 cells had the most connection failures during the outage? & incorrect & 1 & incorrect & 1 & incorrect & 1 \\
Were connection failures elevated on cells behind the affected router during the outage, while the core nodes were also under load? & incorrect & 1 & incorrect & 1 & incorrect & 1 \\
Were connection failures elevated on cells behind the affected router during the outage, even though core node session counts looked normal? & incorrect & 1 & incorrect & 1 & incorrect & 1 \\
How many cells lost availability during the outage on January 2, while the core nodes were also under load? & incorrect & 1 & incorrect & 1 & incorrect & 1 \\
How many cells lost availability during the outage on January 2, compared to the same time the day before when core nodes were healthy? & incorrect & 1 & incorrect & 1 & incorrect & 1 \\
\end{longtable}

\begin{longtable}{>{\raggedright\arraybackslash}p{0.38\textwidth} p{1.7cm} c p{1.7cm} c p{1.7cm} c}
\toprule
\textbf{Query} & \multicolumn{2}{c}{\textbf{Run 1}} & \multicolumn{2}{c}{\textbf{Run 2}} & \multicolumn{2}{c}{\textbf{Run 3}} \\
\cmidrule(lr){2-3} \cmidrule(lr){4-5} \cmidrule(lr){6-7}
 & Class & Label & Class & Label & Class & Label \\
\midrule
\endfirsthead
\toprule
\textbf{Query} & \multicolumn{2}{c}{\textbf{Run 1}} & \multicolumn{2}{c}{\textbf{Run 2}} & \multicolumn{2}{c}{\textbf{Run 3}} \\
\cmidrule(lr){2-3} \cmidrule(lr){4-5} \cmidrule(lr){6-7}
 & Class & Label & Class & Label & Class & Label \\
\midrule
\endhead
\midrule
\multicolumn{7}{r}{\small\textit{(continued)}} \\
\endfoot
\bottomrule
\caption{telecom / PandasAI/Opus4.6 / incident}
\endlastfoot
Did the router that caused the outage have elevated packet loss on January 2 morning? & correct & 1 & incorrect & 2 & incorrect & 2 \\
What was the average latency on the affected router link during the incident? & incorrect & 1 & incorrect & 2 & incorrect & 3 \\
How many transport links were experiencing high packet loss during the outage? & incorrect & 1 & correct & 2 & incorrect & 3 \\
Which transport link had the worst latency during the outage? & correct & 1 & correct & 1 & correct & 1 \\
Were connection failures elevated on cells behind the affected router during the outage? & correct & 1 & incorrect & 2 & incorrect & 2 \\
What was the average connection failure rate on cells behind the affected router during the outage? & incorrect & 1 & incorrect & 2 & incorrect & 3 \\
How many cells lost availability during the outage on January 2? & incorrect & 1 & incorrect & 1 & incorrect & 1 \\
Which 5 cells had the most connection failures during the outage? & correct & 1 & correct & 1 & incorrect & 2 \\
Were connection failures elevated on cells behind the affected router during the outage, while the core nodes were also under load? & incorrect & 1 & correct & 2 & incorrect & 1 \\
Were connection failures elevated on cells behind the affected router during the outage, even though core node session counts looked normal? & correct & 1 & incorrect & 2 & incorrect & 3 \\
How many cells lost availability during the outage on January 2, while the core nodes were also under load? & incorrect & 1 & incorrect & 1 & incorrect & 1 \\
How many cells lost availability during the outage on January 2, compared to the same time the day before when core nodes were healthy? & incorrect & 1 & incorrect & 2 & incorrect & 2 \\
\end{longtable}

\begin{longtable}{>{\raggedright\arraybackslash}p{0.38\textwidth} p{1.7cm} c p{1.7cm} c p{1.7cm} c}
\toprule
\textbf{Query} & \multicolumn{2}{c}{\textbf{Run 1}} & \multicolumn{2}{c}{\textbf{Run 2}} & \multicolumn{2}{c}{\textbf{Run 3}} \\
\cmidrule(lr){2-3} \cmidrule(lr){4-5} \cmidrule(lr){6-7}
 & Class & Label & Class & Label & Class & Label \\
\midrule
\endfirsthead
\toprule
\textbf{Query} & \multicolumn{2}{c}{\textbf{Run 1}} & \multicolumn{2}{c}{\textbf{Run 2}} & \multicolumn{2}{c}{\textbf{Run 3}} \\
\cmidrule(lr){2-3} \cmidrule(lr){4-5} \cmidrule(lr){6-7}
 & Class & Label & Class & Label & Class & Label \\
\midrule
\endhead
\midrule
\multicolumn{7}{r}{\small\textit{(continued)}} \\
\endfoot
\bottomrule
\caption{telecom / Nao/GPT4.1 / incident}
\endlastfoot
Did the router that caused the outage have elevated packet loss on January 2 morning? & correct & 1 & incorrect & 2 & incorrect & 2 \\
What was the average latency on the affected router link during the incident? & incorrect & 1 & runtime\_error & 2 & runtime\_error & 2 \\
How many transport links were experiencing high packet loss during the outage? & incorrect & 1 & runtime\_error & 2 & runtime\_error & 2 \\
Which transport link had the worst latency during the outage? & runtime\_error & 1 & runtime\_error & 1 & incorrect & 2 \\
Were connection failures elevated on cells behind the affected router during the outage? & incorrect & 1 & incorrect & 1 & incorrect & 1 \\
What was the average connection failure rate on cells behind the affected router during the outage? & incorrect & 1 & runtime\_error & 2 & runtime\_error & 2 \\
How many cells lost availability during the outage on January 2? & incorrect & 1 & incorrect & 1 & runtime\_error & 2 \\
Which 5 cells had the most connection failures during the outage? & incorrect & 1 & runtime\_error & 2 & runtime\_error & 2 \\
Were connection failures elevated on cells behind the affected router during the outage, while the core nodes were also under load? & runtime\_error & 1 & runtime\_error & 1 & incorrect & 2 \\
Were connection failures elevated on cells behind the affected router during the outage, even though core node session counts looked normal? & incorrect & 1 & runtime\_error & 2 & runtime\_error & 2 \\
How many cells lost availability during the outage on January 2, while the core nodes were also under load? & runtime\_error & 1 & runtime\_error & 1 & incorrect & 2 \\
How many cells lost availability during the outage on January 2, compared to the same time the day before when core nodes were healthy? & runtime\_error & 1 & incorrect & 2 & runtime\_error & 1 \\
\end{longtable}

\clearpage
\newpage

\section{Agent Responses for Case Studies}
\label{sec:appendix:listings}

\begin{listing}
\begin{minted}[breaklines, frame=single, linenos]{sql}
WITH first_12h AS (
  SELECT *
  FROM `workspace`.`default`.`cell_site_data`
  WHERE `Timestamp` >= (SELECT MIN(`Timestamp`) FROM `workspace`.`default`.`cell_site_data`)
    AND `Timestamp` < (SELECT MIN(`Timestamp`) FROM `workspace`.`default`.`cell_site_data`) + INTERVAL 12 HOURS
    AND `RRC_ConnEstabSucc` IS NOT NULL
)
SELECT AVG(`RRC_ConnEstabSucc`) AS avg_rrc_conn_estab_succ
FROM first_12h;
\end{minted}
\caption{Query: In the first 12 hours of data, what's the average RRC\_ConnEstabSucc?; Agent: Databricks Genie; Dataset: telecom; Query type: stateless, no-incident; Error diagnosis: time interval specified incorrectly in SQL query.}
\label{lst:dbrx_telecom_basic_error}
\end{listing}

\begin{listing}
\begin{minted}[breaklines, frame=single, linenos]{sql}
WITH __cell_site AS (
  SELECT
    rrc_connestabsucc,
    timestamp
  FROM AGENTFUEL.PUBLIC.CELL_SITE
), min_ts AS (
  SELECT
    MIN(timestamp) AS min_timestamp
  FROM __cell_site
)
SELECT
  AVG(rrc_connestabsucc) AS avg_rrc_connestabsucc
FROM __cell_site, min_ts
WHERE
  timestamp >= min_timestamp AND timestamp < DATEADD(HOUR, 12, min_timestamp)
\end{minted}
\caption{Query: In the first 12 hours of data, what's the average RRC\_ConnEstabSucc?; Agent: Snowflake Cortex Analyst; Dataset: telecom; Query type: stateless, no-incident; Error diagnosis: time interval specified incorrectly in SQL query.}
\label{lst:snow_telecom_basic_error}
\end{listing}

\begin{listing}
\begin{minted}[breaklines, frame=single, linenos]{sql}
SELECT AVG(rrc_connestabsucc) AS avg_rrc_connestabsucc
FROM cell_site_data
WHERE timestamp <= (SELECT MIN(timestamp) + interval '12 hours' FROM cell_site_data);
\end{minted}
\caption{Query: In the first 12 hours of data, what's the average RRC\_ConnEstabSucc?; Agent: Nao GPT4.1; Dataset: telecom; Query type: stateless, no-incident; Correct because end timestamp was inclusive.}
\label{lst:nao_telecom_basic_correct}
\end{listing}

\begin{listing}[htp]
\begin{minted}[breaklines, frame=single, linenos]{text}
The column 'RRC\_ConnEstabSucc' does not exist in this dataset. The available columns are: Core\_Node\_ID, MM\_AttachedUEs, SM\_ActivePDUSessions, CPU\_Load, Timestamp. Therefore, the average RRC\_ConnEstabSucc cannot be computed.
\end{minted}
\caption{Example of a GEPA-optimized prompt for a PandasAI O4Mini agent. This prompt showed a 25\% accuracy improvement on the telecom benchmark's incident-specific query set.}
\label{lst:pandasai_basic_error}
\end{listing}

\begin{listing}
\begin{minted}[breaklines, frame=single, linenos]{python}
import pandas as pd
explore_query = """
SELECT DISTINCT page, action 
FROM table_769cb9959208d55e6977f2121826c762
ORDER BY page, action
"""
explore_df = execute_sql_query(explore_query)
count_query = """
WITH sessions_with_cart AS (
    SELECT DISTINCT session_id
    FROM table_769cb9959208d55e6977f2121826c762
    WHERE LOWER(page) LIKE '%
       OR LOWER(action) LIKE '%
       OR LOWER(action) LIKE '%
),
product_views_with_cart AS (
    SELECT t.*
    FROM table_769cb9959208d55e6977f2121826c762 t
    INNER JOIN sessions_with_cart sc ON t.session_id = sc.session_id
    WHERE LOWER(t.page) LIKE '%
       OR LOWER(t.action) LIKE '%
)
SELECT COUNT(*) as product_views_with_cart_count
FROM product_views_with_cart
"""
count_df = execute_sql_query(count_query)
product_view_count = int(
    count_df['product_views_with_cart_count'].iloc[0]
)
result = {'type': 'number', 'value': product_view_count}
\end{minted}
\caption{Query: How many product views occurred while users had an item in their cart?; Agent: PandasAI Sonnet4.6 Dataset: ecommerce; Query type: stateful, no-incident; Error diagnosis: the SQL query should only count view product events that occur after an add to cart and before the next checkout or cart abandoned, within the same session. and it should get the filter value from the data, not use broad matching.}
\label{lst:pandasai_sonnet_ecommerce_stateful_no_inc_error}
\end{listing}

\begin{listing}
\begin{minted}[breaklines, frame=single, linenos]{sql}
SELECT COUNT(*) AS product_views_with_cart
FROM `workspace`.`default`.`ecommerce_sessions_data`
WHERE `action` = 'view_product'
AND `session_id` IN (
    SELECT DISTINCT `session_id`
    FROM `workspace`.`default`.`ecommerce_sessions_data`
    WHERE `action` = 'add_to_cart'
)
AND `action` IS NOT NULL
AND `session_id` IS NOT NULL;
\end{minted}
\caption{Query: How many product views occurred while users had an item in their cart?; Agent: Databricks Genie; Dataset: ecommerce; Query type: stateful, no-incident; Error diagnosis: the SQL query should only count view product events that occur after an add to cart and before the next checkout or cart abandoned, within the same session.}
\label{lst:dbrx_ecommerce_stateful_no_inc_error}
\end{listing}

\begin{listing}
\begin{minted}[breaklines, frame=single, linenos]{sql}
WITH __ecommerce_sessions_data AS (
  SELECT
    action,
    page,
    session_id
  FROM AGENTFUEL.PUBLIC.ECOMMERCE_SESSIONS_DATA
), cart_sessions AS (
  SELECT DISTINCT
    session_id
  FROM __ecommerce_sessions_data
  WHERE
    action = 'add_to_cart'
)
SELECT
  COUNT(*) AS product_views_with_cart
FROM __ecommerce_sessions_data AS s
JOIN cart_sessions AS c
  ON s.session_id = c.session_id
WHERE
  s.page = 'product';
\end{minted}
\caption{Query: How many product views occurred while users had an item in their cart?; Agent: Snowflake Cortex Analyst; Dataset: ecommerce; Query type: stateful, no-incident; Error diagnosis: the SQL query should only count view product events that occur after an add to cart and before the next checkout or cart abandoned, within the same session.}
\label{lst:snow_ecommerce_stateful_no_inc_error}
\end{listing}

\begin{listing}
\begin{minted}[breaklines, frame=single, linenos]{python}
import pandas as pd
sql_query = """
SELECT 
    device_id,
    device_state,
    timestamp
FROM table_d432f863c608f24c8942f4c714947b86
WHERE firmware_version = 'v2.0'
  AND device_state IN ('warning', 'critical', 'maintenance', 'operational')
ORDER BY device_id, timestamp
"""
df = execute_sql_query(sql_query)

def check_sequence(states):
    """Check if the sequence warning -> critical -> maintenance -> operational exists in order"""
    target_sequence = ['warning', 'critical', 'maintenance', 'operational']
    deduped = []
    for state in states:
        if not deduped or deduped[-1] != state:
            deduped.append(state)
    seq_idx = 0
    for state in deduped:
        if seq_idx < len(target_sequence) and state == target_sequence[seq_idx]:
            seq_idx += 1
        if seq_idx == len(target_sequence):
            return True
    return False

matching_devices = []
for device_id, group in df.groupby('device_id'):
    states = (
        group
        .sort_values('timestamp')
        ['device_state']
        .tolist()
    )
    if check_sequence(states):
        matching_devices.append(device_id)
count = len(matching_devices)
print(f'Number of v2.0 sensors that went through warning -> critical -> maintenance -> operational: {count}')
\end{minted}
\caption{Query: Were any v2.0 sensors in warning, critical, maintenance, and then operational?; Agent: PandasAI Sonnet4.6; Dataset: iot; Query type: stateful, no-incident; Correct answer.}
\label{lst:pandasai_sonnet_iot_stateful_correct}
\end{listing}

\begin{listing}
\begin{minted}[breaklines, frame=single, linenos]{sql}
SELECT DISTINCT t1.`device_id`
FROM
  (SELECT `device_id`,
          `device_state`,
          `timestamp`,
          ROW_NUMBER() OVER (PARTITION BY `device_id`
                             ORDER BY `timestamp`) AS rn
   FROM `workspace`.`default`.`iot_device_data`
   WHERE `firmware_version` = 'v2.0') t1
JOIN
  (SELECT `device_id`,
          `device_state`,
          `timestamp`,
          ROW_NUMBER() OVER (PARTITION BY `device_id`
                             ORDER BY `timestamp`) AS rn
   FROM `workspace`.`default`.`iot_device_data`
   WHERE `firmware_version` = 'v2.0') t2 ON t1.`device_id` = t2.`device_id`
AND t1.rn = t2.rn - 1
JOIN
  (SELECT `device_id`,
          `device_state`,
          `timestamp`,
          ROW_NUMBER() OVER (PARTITION BY `device_id`
                             ORDER BY `timestamp`) AS rn
   FROM `workspace`.`default`.`iot_device_data`
   WHERE `firmware_version` = 'v2.0') t3 ON t1.`device_id` = t3.`device_id`
AND t1.rn = t3.rn - 2
JOIN
  (SELECT `device_id`,
          `device_state`,
          `timestamp`,
          ROW_NUMBER() OVER (PARTITION BY `device_id`
                             ORDER BY `timestamp`) AS rn
   FROM `workspace`.`default`.`iot_device_data`
   WHERE `firmware_version` = 'v2.0') t4 ON t1.`device_id` = t4.`device_id`
AND t1.rn = t4.rn - 3
WHERE t1.`device_state` = 'operational'
  AND t2.`device_state` = 'maintenance'
  AND t3.`device_state` = 'critical'
  AND t4.`device_state` = 'warning'
\end{minted}
\caption{Query: Were any v2.0 sensors in warning, critical, maintenance, and then operational?; Agent: Databricks Genie; Dataset: iot; Query type: stateful, no-incident; Error diagnosis: Exact sequence matching instead of subsequence match, order of sequence is reversed.}
\label{lst:dbrx_iot_stateful_no_inc_error}
\end{listing}

\begin{listing}
\begin{minted}[breaklines, frame=single, linenos]{sql}
WITH __iot_device_data AS (
  SELECT
    device_id,
    device_state,
    firmware_version,
    timestamp
  FROM AGENTFUEL.PUBLIC.IOT_DEVICE_DATA
), state_changes AS (
  SELECT
    device_id,
    timestamp,
    device_state,
    firmware_version,
    LAG(device_state, 1) OVER (PARTITION BY device_id ORDER BY timestamp) AS prev_state,
    LAG(device_state, 2) OVER (PARTITION BY device_id ORDER BY timestamp) AS prev_state2,
    LAG(device_state, 3) OVER (PARTITION BY device_id ORDER BY timestamp) AS prev_state3
  FROM __iot_device_data
  WHERE
    firmware_version = 'v2.0'
)
SELECT DISTINCT
  device_id
FROM state_changes
WHERE
  device_state = 'operational'
  AND prev_state = 'maintenance'
  AND prev_state2 = 'critical'
  AND prev_state3 = 'warning'
ORDER BY
  device_id DESC NULLS LAST;
\end{minted}
\caption{Query: Were any v2.0 sensors in warning, critical, maintenance, and then operational?; Agent: Snowflake Cortex Analyst; Dataset: iot; Query type: stateful, no-incident; Error diagnosis: exact sequence match instead of subsequence.}
\label{lst:snow_iot_stateful_no_inc_error}
\end{listing}

\begin{listing}
\begin{minted}[breaklines, frame=single, linenos]{sql}
SELECT AVG(latency_ms) AS avg_latency
FROM public.transport_link_with_inc_data
WHERE device_id = 'RTR_004'
  AND interface_id = 'eth1';
\end{minted}
\caption{Query: What was the average latency on the affected router link during the incident?; Agent: Nao GPT4.1; Dataset: telecom; Query type: stateless, incident; Error diagnosis: no exploration for incident detection, random router picked.}
\label{lst:nao_telecom_stateless_inc_error}
\end{listing}

\begin{listing}
\begin{minted}[breaklines, frame=single, linenos]{sql}
SELECT AVG(Latency_ms) AS avg_latency
FROM `workspace`.`default`.`transport_link_with_inc_data`
WHERE Latency_ms IS NOT NULL;
\end{minted}
\caption{Query: What was the average latency on the affected router link during the incident?; Agent: Databricks Genie; Dataset: telecom; Query type: stateless, incident; Error diagnosis: no exploration for incident detection, average over entire data.}
\label{lst:dbrx_telecom_stateless_inc_error}
\end{listing}

\begin{listing}
\begin{minted}[breaklines, frame=single, linenos]{python}
import pandas as pd

transport_stats = execute_sql_query(
    """
    SELECT 
        Device_ID,
        Interface_ID,
        AVG(Packet_Loss_Percent) as avg_packet_loss,
        MAX(Packet_Loss_Percent) as max_packet_loss,
        AVG(Latency_ms) as avg_latency,
        MAX(Latency_ms) as max_latency,
        STDDEV(Packet_Loss_Percent) as std_packet_loss,
        STDDEV(Latency_ms) as std_latency
    FROM table_1b020433bc09869c9bfed7f8dadacca6
    GROUP BY Device_ID, Interface_ID
    ORDER BY max_packet_loss DESC
"""
)
print("Transport link stats by device/interface:")
print(transport_stats)
incident_data = execute_sql_query(
    """
    WITH stats AS (
        SELECT 
            AVG(Packet_Loss_Percent) as mean_loss,
            STDDEV(Packet_Loss_Percent) as std_loss,
            AVG(Latency_ms) as mean_latency,
            STDDEV(Latency_ms) as std_latency
        FROM table_1b020433bc09869c9bfed7f8dadacca6
    ),
    anomalies AS (
        SELECT 
            t.*,
            s.mean_loss,
            s.std_loss,
            s.mean_latency,
            s.std_latency
        FROM table_1b020433bc09869c9bfed7f8dadacca6 t
        CROSS JOIN stats s
        WHERE t.Packet_Loss_Percent > (s.mean_loss + 2 * s.std_loss)
           OR t.Latency_ms > (s.mean_latency + 2 * s.std_latency)
    )
    SELECT * FROM anomalies
    ORDER BY Timestamp
"""
)
print("""Incident periods (anomalous router data):""")
print(incident_data)
print(f"Number of incident records: {len(incident_data)}")
affected_link = execute_sql_query(
    """
    WITH stats AS (
        SELECT 
            AVG(Packet_Loss_Percent) as mean_loss,
            STDDEV(Packet_Loss_Percent) as std_loss,
            AVG(Latency_ms) as mean_latency,
            STDDEV(Latency_ms) as std_latency
        FROM table_1b020433bc09869c9bfed7f8dadacca6
    ),
    incident_records AS (
        SELECT 
            t.Device_ID,
            t.Interface_ID,
            t.Latency_ms,
            t.Packet_Loss_Percent,
            t.Timestamp
        FROM table_1b020433bc09869c9bfed7f8dadacca6 t
        CROSS JOIN stats s
        WHERE t.Packet_Loss_Percent > (s.mean_loss + 2 * s.std_loss)
           OR t.Latency_ms > (s.mean_latency + 2 * s.std_latency)
    )
    SELECT 
        Device_ID,
        Interface_ID,
        COUNT(*) as incident_count,
        AVG(Latency_ms) as avg_latency_during_incident,
        AVG(Packet_Loss_Percent) as avg_packet_loss_during_incident
    FROM incident_records
    GROUP BY Device_ID, Interface_ID
    ORDER BY incident_count DESC
"""
)
print(
    """
Affected router links during incident:"""
)
print(affected_link)
avg_latency_incident = execute_sql_query(
    """
    WITH stats AS (
        SELECT 
            AVG(Packet_Loss_Percent) as mean_loss,
            STDDEV(Packet_Loss_Percent) as std_loss,
            AVG(Latency_ms) as mean_latency,
            STDDEV(Latency_ms) as std_latency
        FROM table_1b020433bc09869c9bfed7f8dadacca6
    ),
    incident_records AS (
        SELECT 
            t.Device_ID,
            t.Interface_ID,
            t.Latency_ms,
            t.Packet_Loss_Percent
        FROM table_1b020433bc09869c9bfed7f8dadacca6 t
        CROSS JOIN stats s
        WHERE t.Packet_Loss_Percent > (s.mean_loss + 2 * s.std_loss)
           OR t.Latency_ms > (s.mean_latency + 2 * s.std_latency)
    )
    SELECT 
        Device_ID,
        Interface_ID,
        ROUND(AVG(Latency_ms), 2) as avg_latency_ms
    FROM incident_records
    GROUP BY Device_ID, Interface_ID
    ORDER BY avg_latency_ms DESC
    LIMIT 1
"""
)
print(
    """
Average latency on affected router link during incident:"""
)
print(avg_latency_incident)
if len(avg_latency_incident) > 0:
    device_id = avg_latency_incident["Device_ID"].iloc[0]
    interface_id = avg_latency_incident["Interface_ID"].iloc[0]
    avg_latency = avg_latency_incident["avg_latency_ms"].iloc[0]
    result = {
        "type": "string",
        "value": f"During the incident, the average latency on the affected router link ({device_id} / {interface_id}) was {avg_latency:.2f} ms.",
    }
else:
    overall_latency = execute_sql_query(
        """
        SELECT 
            Device_ID,
            Interface_ID,
            ROUND(AVG(Latency_ms), 2) as avg_latency_ms
        FROM table_1b020433bc09869c9bfed7f8dadacca6
        GROUP BY Device_ID, Interface_ID
        ORDER BY avg_latency_ms DESC
        LIMIT 1
    """
    )
    device_id = overall_latency["Device_ID"].iloc[0]
    interface_id = overall_latency["Interface_ID"].iloc[0]
    avg_latency = overall_latency["avg_latency_ms"].iloc[0]
    result = {
        "type": "string",
        "value": f"No clear incident detected. The overall average latency on the highest-latency router link ({device_id} / {interface_id}) was {avg_latency:.2f} ms.",
    }
print(result)
\end{minted}
\caption{Query: What was the average latency on the affected router link during the incident?; Agent: PandasAI Sonnet4.6; Dataset: telecom; Query type: stateless, incident; Error diagnosis: dataset exploration done for incident detection, but incorrect logic (baseline is not entity-specific i.e. logic should check if an incident occurred for an entity by comparing its own average behavior, not globally).}
\label{lst:pandasai_sonnet_telecom_stateless_inc_error}
\end{listing}

\begin{listing}
\begin{minted}[breaklines, frame=single, linenos]{sql}
SELECT COUNT(DISTINCT c.cell_id) AS lost_cells
FROM cell_site_with_inc_data c
JOIN core_node_with_inc_data n
  ON c.timestamp = n.timestamp
WHERE DATE(c.timestamp) = '2026-01-02'
  AND c.cell_availability < 99.0
  AND n.cpu_load > 50.0;
\end{minted}
\caption{Query: How many cells lost availability during the outage on January 2, while the core nodes were also under load?; Agent: Nao GPT4.1; Dataset: telecom; Query type: stateful, incident; Error diagnosis: thresholds, baseline data slice to compare with are not data-aware.}
\label{lst:nao_telecom_stateful_inc_error}
\end{listing}

\begin{listing}
\begin{minted}[breaklines, frame=single, linenos]{python}
import pandas as pd
query = """
WITH cell_outage AS (
    SELECT 
        Cell_ID,
        Timestamp,
        Cell_Availability,
        Base_Station_ID
    FROM table_a828d9ac66625631ed33842c652a9e27
    WHERE Timestamp >= '2026-01-02T00:00:00' 
      AND Timestamp < '2026-01-03T00:00:00'
      AND Cell_Availability < 99.0
),
core_load AS (
    SELECT 
        Core_Node_ID,
        Timestamp,
        CPU_Load
    FROM table_0be82d4a8d18415e6d6e3b8c90d346d0
    WHERE Timestamp >= '2026-01-02T00:00:00' 
      AND Timestamp < '2026-01-03T00:00:00'
      AND CPU_Load > 70.0
),
cells_during_core_load AS (
    SELECT DISTINCT c.Cell_ID
    FROM cell_outage c
    INNER JOIN core_load cl
        ON c.Timestamp = cl.Timestamp
)
SELECT COUNT(*) as num_cells
FROM cells_during_core_load
"""
df_result = execute_sql_query(query)
num_cells = df_result['num_cells'].iloc[0]
if num_cells == 0:
    explore_query = """
    SELECT 
        MIN(Cell_Availability) as min_avail,
        AVG(Cell_Availability) as avg_avail,
        MAX(Cell_Availability) as max_avail,
        PERCENTILE_CONT(0.05) WITHIN GROUP (ORDER BY Cell_Availability) as p5_avail
    FROM table_a828d9ac66625631ed33842c652a9e27
    WHERE Timestamp >= '2026-01-02T00:00:00' 
      AND Timestamp < '2026-01-03T00:00:00'
    """
    df_explore = execute_sql_query(explore_query)
    explore_core = """
    SELECT 
        MIN(CPU_Load) as min_cpu,
        AVG(CPU_Load) as avg_cpu,
        MAX(CPU_Load) as max_cpu,
        PERCENTILE_CONT(0.75) WITHIN GROUP (ORDER BY CPU_Load) as p75_cpu
    FROM table_0be82d4a8d18415e6d6e3b8c90d346d0
    WHERE Timestamp >= '2026-01-02T00:00:00' 
      AND Timestamp < '2026-01-03T00:00:00'
    """
    df_core = execute_sql_query(explore_core)
    avg_cpu = df_core['avg_cpu'].iloc[0]
    avg_avail = df_explore['avg_avail'].iloc[0]
    refined_query = f"""
    WITH cell_outage AS (
        SELECT DISTINCT Cell_ID, Timestamp
        FROM table_a828d9ac66625631ed33842c652a9e27
        WHERE Timestamp >= '2026-01-02T00:00:00' 
          AND Timestamp < '2026-01-03T00:00:00'
          AND Cell_Availability < {avg_avail}
    ),
    core_load AS (
        SELECT DISTINCT Timestamp
        FROM table_0be82d4a8d18415e6d6e3b8c90d346d0
        WHERE Timestamp >= '2026-01-02T00:00:00' 
          AND Timestamp < '2026-01-03T00:00:00'
          AND CPU_Load > {avg_cpu}
    ),
    cells_during_core_load AS (
        SELECT DISTINCT c.Cell_ID
        FROM cell_outage c
        INNER JOIN core_load cl
            ON c.Timestamp = cl.Timestamp
    )
    SELECT COUNT(*) as num_cells
    FROM cells_during_core_load
    """
    df_result2 = execute_sql_query(refined_query)
    num_cells = df_result2['num_cells'].iloc[0]
result = {'type': 'string', 'value': f'{num_cells} cells lost availability during the outage on January 2 while the core nodes were also under load.'}
\end{minted}
\caption{query: How many cells lost availability during the outage on January 2, while the core nodes were also under load? Agent: pandasai/opus4.6 Dataset: telecom query: stateful, incident. Error diagnosis: thresholds, baseline data slice to compare with are not data-aware}
\label{lst:pandasai_opus_telecom_stateful_inc_error}
\end{listing}

\begin{listing}
\begin{minted}[breaklines, frame=single, linenos]{sql}
SELECT COUNT(DISTINCT c.`Cell_ID`) AS cells_lost_availability
FROM `workspace`.`default`.`cell_site_with_inc_data` c
JOIN `workspace`.`default`.`core_node_with_inc_data` n ON c.`Timestamp` = n.`Timestamp`
WHERE DATE(c.`Timestamp`) = '2026-01-02'
  AND DATE(n.`Timestamp`) = '2026-01-02'
  AND c.`Cell_Availability` < 100
  AND n.`CPU_Load` > 0
  AND c.`Cell_ID` IS NOT NULL
  AND c.`Cell_Availability` IS NOT NULL
  AND n.`CPU_Load` IS NOT NULL
\end{minted}
\caption{Query: How many cells lost availability during the outage on January 2, while the core nodes were also under load?; Agent: Databricks Genie; Dataset: telecom; Query type: stateful, incident; Error diagnosis: thresholds, baseline data slice to compare with are not data-aware.}
\label{lst:dbrx_telecom_stateful_inc_error}
\end{listing}

\twocolumn

\clearpage \newpage

\section{Additional  Query Templates}
\label{sec:appendix:morequeries}

\begin{table*}[h]
\renewcommand{\arraystretch}{1.4}
\caption{More stateful query templates.
  $\entity$: entity; $\mystate,\mystate'$: states in $\StateSet$;
  $\QueryWindow$: window; $\EventPred{}$: event predicate.}
\label{tab:stateful2}
\smallskip
\begin{tabularx}{\textwidth}{
    p{0.1\textwidth}
    p{0.2\textwidth}
    >{\raggedright\arraybackslash}X
    p{0.4\textwidth}
  }
\toprule
 \textbf{Name} & \textbf{Formal Expression} & \textbf{Description} \\
\midrule

State reached &
$\StateReached{\entity}{\mystate}{\QueryWindow} = \mathbf{1}\!\left[\StateOcc{\entity}{\mystate}{\QueryWindow} \neq \emptyset\right]$ &
Boolean: did $\entity$ enter $\mystate$ at least once in $\QueryWindow$? Aggregated to yield entities-reached count and average first-entry time. \\[4pt]

Count events in state &
$\CountInState{\entity}{\mystate}{\EventPred{}}{\QueryWindow} = \bigl|\{\ttime \in \MeasSet{\entity}\cap\QueryWindow : \EventPred{}(\MeasPayload{\entity}{\ttime}),\; \ActiveState{\entity}{\ttime}=\mystate\}\bigr|$ &
Events matching $\EventPred{}$ that occurred while $\entity$ was in $\mystate$; conditions event counts on behavioural context. \\[4pt]

State duration &
$\StateDur{\entity}{\mystate}{\QueryWindow} = \displaystyle\sum_{[t^{\mathrm{in}},t^{\mathrm{out}})\,\in\,\StateOcc{\entity}{\mystate}{\QueryWindow}} (t^{\mathrm{out}} - t^{\mathrm{in}})$ &
Total sojourn time in $\mystate$ within $\QueryWindow$.  Aggregated as mean, median, or $\Percentile{p}{}$ across entities; supports exit-event stratification. \\[4pt]

First passage time &
$\FirstPassage{\entity}{\mystate}{\ttime_0} = \inf\!\left\{\ttime \geq \ttime_0 : \ActiveState{\entity}{\ttime} = \mystate\right\} - \ttime_0$ &
Elapsed time from reference $\ttime_0$ (e.g.\ entity creation) until the first entry into $\mystate$. \\[4pt]

Transition matrix &
$\TransMatrix{\EntitySet'}{\QueryWindow}_{(\mystate,\mystate')} = \displaystyle\sum_{\entity\in\EntitySet'}\bigl|\{j : \ActiveState{\entity}{\ttime_j^{\mathrm{out}}}=\mystate,\;\ActiveState{\entity}{\ttime_{j+1}^{\mathrm{in}}}=\mystate'\}\bigr|$ &
Counts of all $\mystate\!\to\!\mystate'$ transitions; row-normalised to give empirical probabilities $\TransProb{\entity}{\mystate}{\mystate'}$. \\
Common trajectory paths &
$\CommonPaths{\EntitySet'}{\QueryWindow} = \mathrm{top}\text{-}n\;\text{sequences}\;\langle\state_1,\ldots,\state_\ell\rangle$ &
Most frequently observed ordered state sequences (up to length $\ell$) across entities; reveals dominant behavioural pathways. \\[4pt]

Loop / revisit count &
$\LoopCount{\entity}{\mystate}{\QueryWindow} = \bigl|\StateOcc{\entity}{\mystate}{\QueryWindow}\bigr| - \mathbf{1}[\StateOcc{\entity}{\mystate}{\QueryWindow}\neq\emptyset]$ &
Number of times $\entity$ re-entered $\mystate$ after having previously exited; detects oscillating or cycling behaviour. \\[4pt]

Multi-state occupancy &
$\OccupancyDist{\entity}{\StateSet}{\QueryWindow}_{\mystate} = \dfrac{\StateDur{\entity}{\mystate}{\QueryWindow}}{\displaystyle\sum_{\mystate'\in\StateSet}\StateDur{\entity}{\mystate'}{\QueryWindow}}$ &
Fraction of observed time spent in each state; summarises the full behavioural profile of an entity across the state space. \\[4pt]

KPI conditioned on state &
$\KPI{f}{\entity}{\QueryWindow \cap \StateOcc{\entity}{\mystate}{\QueryWindow}}$ &
Evaluates any stateless KPI $f \in \KPISet$ restricted to sub-windows during which $\entity$ occupied $\mystate$; bridges the two query families. \\

\bottomrule
\end{tabularx}
\end{table*}

\end{document}